\documentclass[compsoc,journal]{IEEEtran}
\usepackage{array}     
\usepackage{tabularx}  
\usepackage{ragged2e}  
\usepackage{adjustbox}
\usepackage{multirow} 
\usepackage{amsmath,amsfonts}
\usepackage{hyperref}
\usepackage{algorithm}
\usepackage{amsmath}

\usepackage[caption=false,font=normalsize,labelfont=sf,textfont=sf]{subfig}
\usepackage{textcomp}
\usepackage{stfloats}
\usepackage{url}
\usepackage{verbatim}
\usepackage{graphicx}
\usepackage{cite}
\usepackage{booktabs} 
\usepackage{tikz}
\usepackage{booktabs}
\usepackage{multirow}
\usepackage{siunitx}
\usepackage{graphicx} 
\usepackage{tabularx}   
\usepackage{ragged2e}   
\usepackage{makecell}   

\usepackage{booktabs}
\usepackage{multirow}
\usepackage{graphicx}
\usepackage{xcolor}
\usepackage{colortbl} 
\usepackage{amsmath}
\usepackage{algorithm}
\usepackage{algpseudocode}
\usepackage{amsmath}

\definecolor{headergray}{gray}{0.92}      
\definecolor{oursbg}{RGB}{236, 245, 255}  

\definecolor{posdelta}{RGB}{0, 130, 60}   
\definecolor{negdelta}{gray}{0.65}        

\usepackage[compatibility=false]{caption}
\newcolumntype{C}{>{\centering\arraybackslash}p{0.8cm}}

\definecolor{headergray}{gray}{0.92}
\definecolor{categorygray}{gray}{0.96}  
\definecolor{oursbg}{RGB}{236, 245, 255}

\usepackage[table]{xcolor} 

\newcolumntype{Y}{>{\RaggedRight\arraybackslash}X}
\newcolumntype{Z}{>{\Centering\arraybackslash}X}

\begin{document}

\title{U-MASK: User-adaptive Spatio-Temporal Masking for Personalized Mobile AI Applications}

\author{
Shiyuan Zhang,
Yilai Liu,
Yuwei Du,
Ruoxuan Yang,
Dong In Kim,~\IEEEmembership{Fellow,~IEEE},
and Hongyang Du
\thanks{
S. Zhang, Y. Liu, R. Yang, and H. Du are with the Department of Electrical and Electronic Engineering, The University of Hong Kong, Pok Fu Lam, Hong Kong SAR, China
(e-mail: \{shiyuanzhang, yilai\_liu\}@connect.hku.hk; rosheenanea@outlook.com; duhy@eee.hku.hk).
}
\thanks{
Y. Du is with the Department of Electronic Engineering, Tsinghua University, Beijing, China.
(e-mail: duyw23@mails.tsinghua.edu.cn).
}
\thanks{
D. I. Kim is with the Department of Electrical and Computer Engineering, Sungkyunkwan
University, Suwon 16419, South Korea (e-mail: dongin@skku.edu).
}
}

\maketitle

\begin{abstract}
Personalized mobile artificial intelligence applications are widely deployed, yet they are expected to infer user behavior from sparse and irregular histories under a continuously evolving spatio-temporal context. This setting induces a fundamental tension among three requirements, i.e., immediacy to adapt to recent behavior, stability to resist transient noise, and generalization to support long-horizon prediction and cold-start users. 
Most existing approaches satisfy at most two of these requirements, resulting in an inherent impossibility triangle in data-scarce, non-stationary personalization. 
To address this challenge, we model mobile behavior as a partially observed spatio-temporal tensor and unify short-term adaptation, long-horizon forecasting, and cold-start recommendation as a conditional completion problem, where a user- and task-specific mask specifies which coordinates are treated as evidence. 
We propose U-MASK, a user-adaptive spatio-temporal masking method that allocates evidence budgets based on user reliability and task sensitivity. To enable mask generation under sparse observations, U-MASK learns a compact, task-agnostic user representation from app and location histories via U-SCOPE, which serves as the sole semantic conditioning signal. A shared diffusion transformer then performs mask-guided generative completion while preserving observed evidence, so personalization and task differentiation are governed entirely by the mask and the user representation. Experiments on real-world mobile datasets demonstrate consistent improvements over state-of-the-art methods across short-term prediction, long-horizon forecasting, and cold-start settings, with the largest gains under severe data sparsity. The code and dataset will be available at \href{https://github.com/NICE-HKU/U-MASK} {https://github.com/NICE-HKU/U-MASK}.
\end{abstract}


\section{Introduction}\label{sec:introduction}
\IEEEPARstart{M}{obile} Artificial Intelligence (AI) applications (apps) have evolved from optional enhancements into central components of everyday digital experiences as device intelligence and app ecosystems continue to advance. 
In the first half of 2025 alone, AI-driven mobile apps surpassed $7.5$ billion global downloads, with a $52\%$ annual growth rate~\cite{sensor2025state}. 
This rapid growth reflects increasing demand for personalized assistants productivity tools, and generative platforms that rely on precise user behavior modeling and strong contextual awareness.
Unlike conventional mobile systems, these AI-driven apps require accurate inference of user behavior and the ability to adapt such inference to dynamic user contexts, behavioral intent, and varying environmental conditions~\cite{wang2016data, lu2020user}.

Although this shift has directed increasing attention toward improving system-level AI service quality, the modeling of users' mobile environments and AI usage behavior has progressed much more slowly than the community's emphasis on solely optimizing model performance. 
Mobile AI apps rely on continuous data exchange, on-device inference, and cross-device service continuity, all of which are largely shaped by network variability as the underlying computing and data substrate. As a result, user interactions and computation patterns vary significantly across individuals, locations, and temporal contexts, leading to pronounced heterogeneity in mobile traffic~\cite{xu2016understanding, naboulsi2015large}. Accurate modeling must capture user-specific spatio-temporal regularities together with the contextual factors that govern them. These factors affect not only traffic demand but also the timeliness and reliability of AI app. For instance, data demand may peak consistently during a user's daily commute in dense urban areas, whereas another user exhibits stable media consumption at home in the evening. Capturing such individual-level variation is critical for designing user-centric systems that anticipate user needs rather than responding after the fact~\cite{laghari2024review, zhang2024netdiff}.

Despite growing interest in personalized mobile AI, existing approaches that incorporate behavioral and contextual information typically target a single narrowly scoped task, which limits their ability to support the diverse requirements of real-world app. Specifically, a dominant line of work focuses on aggregate-level traffic forecasting, modeling collective behavior across user groups or base stations to predict overall traffic volume or bandwidth demand~\cite{chai2024knowledge, hui2023large, zhou2024graph, liu2024spatial}. By design, these methods discard individual heterogeneity and cannot support user-specific service adaptation. To recover personalization, another body of research formulates app usage modeling as a sequential next-item prediction problem based on historical interactions~\cite{zhang2021dual, shi2020spatial}. While effective for short-term repetition, this paradigm breaks down for cold-start users with sparse records and cannot recommend new app beyond past usage. Recent efforts enrich behavior modeling with contextual signals such as time, location, or human activity, yet most process these modalities in isolation and fail to capture their joint spatio-temporal dependencies~\cite{huang2025appgen, duan2025attrappgen, wang2021app2vec}. Even recent advanced architectures lack explicit spatial grounding~\cite{meng2025tuning, gong2025sttf} or rely on coarse spatial discretization that smooths out local variation~\cite{li2021spatiotemporal, zhang2021dual}.

\begin{figure}[t]
\centering
\includegraphics[width=0.47\textwidth]{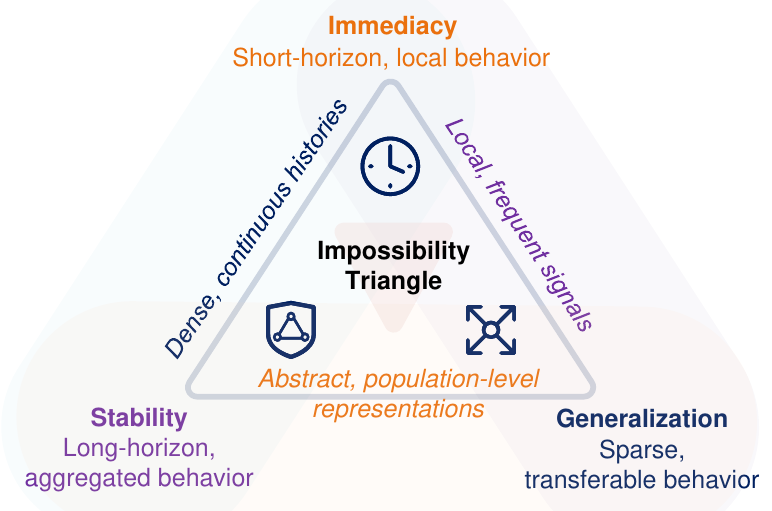}
\caption{The impossibility triangle of personalized mobile AI app usage behavior modeling. Each edge represents a shared modeling assumption that enables the two adjacent objectives while structurally excluding the third.}
\label{fig:triangle}
\end{figure}

These limitations expose a fundamental gap in existing personalized mobile AI models as illustrated in Fig.~\ref{fig:triangle}. This gap is structural rather than empirical and appears as an inherent tension among three requirements in modeling how people use personalized mobile AI:
\begin{itemize}
    \item \textbf{Immediacy}: accurate inference from recent and local behavioral context, which corresponds to \emph{short-term prediction} for time-critical network management and service adaptation;
    \item \textbf{Stability}: extracting periodic structure and aggregated behavioral pattern, which underpins \emph{long-term forecasting} for planning-oriented decisions such as resource provisioning and service scheduling;
    \item \textbf{Generalization}: maintaining reliable inference under extreme sparsity or distribution shift, which is required by \emph{cold-start recommendation} when historical traces are limited or newly emerging contexts appear.
\end{itemize}
This tension arises because each requirement imposes different demands on the use of behavioral evidence.  Most existing methods address this tension implicitly by fixing the prediction target and the conditioning structure in advance, thereby forcing a trade-off among these requirements.  Approaches that aim to capture both {\textbf{immediacy}} and {\textbf{stability}} typically rely on dense historical data through sequence-based or spatio-temporal modeling~\cite{zhang2021dual, shi2020spatial, li2021spatiotemporal}. These models rely on continuous temporal coverage to estimate short-term dynamics and long-term trends simultaneously. However, their performance degrades sharply when user histories are sparse, limiting generalization.  Methods that prioritize {\textbf{immediacy}} and {\textbf{generalization}} instead constrain inference to highly local or frequent patterns, such as short-horizon repetition or popularity-based signals. While this reduces data requirements, it prevents the model from accumulating and preserving long-term behavioral structure.  Conversely, models designed to achieve {\textbf{stability}} and {\textbf{generalization}} abstract behavior at the aggregate or profile level~\cite{chai2024knowledge, hui2023large, zhou2024graph}. This abstraction smooths out individual variation and makes the model insensitive to time-critical user context. As a result, existing solutions typically satisfy at most two corners of this ``impossibility triangle'' while inevitably compromising the third.

In practice, this fragmentation appears how the requirements of the impossibility triangle are mapped to modeling tasks. 
Short-term forecasting, long-term trend prediction, and cold-start recommendation under sparse histories are cast as separate problems and addressed with different objectives. Yet these tasks operate on the same underlying spatio-temporal behavioral process and differ primarily in how available observations are partitioned into evidence and targets. 
In all cases, the model reasons over partial observations of a shared process, with the distinction determined by which regions are treated as context and which are inferred. This observation motivates a reformulation of personalized mobile behavior modeling as a conditional completion problem over sparse spatio-temporal data, where different task regimes correspond to different completion patterns imposed on a shared behavioral representation rather than fundamentally different learning problems.

From this perspective, masking becomes a structural mechanism for defining inference rather than an auxiliary training technique. Masked modeling provides a general formulation of structured prediction as conditional completion, in which models learn data regularities by reconstructing partially observed inputs. This paradigm was first established in natural language processing through masked language modeling~\cite{devlin2019bert}, and later extended to vision with masked autoencoders such as MAE~\cite{he2022masked} and BeiT~\cite{bao2021beit}. Recent spatio-temporal extensions, including VideoMAE~\cite{tong2022videomae} and UMT~\cite{li2023unmasked}, further show that masking strategies encode inductive bias by shaping how temporal continuity and cross-modal dependencies are captured. In personalized mobile behavioral data, such as app usage traces coupled with spatio-temporal context, a mask specifies which regions of the behavior tensor are treated as evidence and which are inferred, thereby controlling the temporal horizon, spatial scope, and effective information available for inference. This shifts the core challenge from designing task-specific predictors to learning adaptive mechanisms that define the spatio-temporal support of inference itself.

Therefore, we reframe personalized mobile behavior modeling as a conditional completion problem over sparse spatio-temporal observations, where inference is explicitly defined by which regions are observed and which are reconstructed. Under this formulation, the model must infer missing behavioral structure from highly incomplete and unevenly distributed data while remaining sensitive to individual routines and contextual intent, construct meaningful user semantics from raw app usage and spatio-temporal traces, and accommodate heterogeneous inference horizons ranging from short-term adaptation to long-term regularity without handcrafted context definitions. Our key contributions are summarized as follows:
\begin{itemize}
    \item We propose \textbf{U-MASK}, a user- and task-specific spatio-temporal masking mechanism that serves as the modeling abstraction for personalized mobile AI. By treating the mask as the primary object that defines both the task target and its conditioning context, U-MASK formulates personalized behavior modeling as a unified completion problem and resolves the structural tension among immediacy, stability, and generalization through mask configuration.
    
    \item We develop \textbf{U-SCOPE}, a semantic user profiling framework under sparse and heterogeneous observations for constructing informative spatio-temporal masks. U-SCOPE distills raw app usage and location telemetry into stable intent-level representations by capturing recurring behavioral co-occurrence patterns, which guide the spatial scope and temporal sensitivity of user-adaptive masking and enable robust personalization in data-scarce scenarios.
    
    \item We instantiate the U-MASK formulation with a conditional diffusion-based generative architecture. This diffusion transformer realization supports diverse personalized task objectives, including short-term behavior prediction, long-term forecasting, and cold-start recommendation, within a single generative process. Extensive experiments on real-world mobile datasets demonstrate consistent improvements over state-of-the-art methods in both predictive accuracy and behavioral realism, particularly under dynamic contexts and sparse user histories.
\end{itemize}

The remainder of this paper is organized as follows. Section~\ref{sec:related} reviews related work and identifies the absence of a user-adaptive masking mechanism. Section~\ref{sec:overview} presents the overall framework, which formulates personalized mobile AI as a mask-conditioned generative completion problem. 
Section~\ref{sec:umask} introduces the core algorithmic contribution of U-MASK. 
Section~\ref{sec:uscope} details U-SCOPE, the semantic profiling engine that generates high-fidelity user embeddings from sparse telemetry to condition the mask. 
Section~\ref{sec:diffusion} introduces the shared Diffusion Transformer (DiT) backbone.
Section~\ref{sec:exp} presents extensive experiments validating the effectiveness of our approach.
Section~\ref{con} concludes by summarizing the U-MASK framework, which enables universal, mask-driven personalization, and highlights its empirical validation and future potential.

\section{Related work}\label{sec:related}
In this section, we review related work on user behavior modeling for mobile AI apps, masked modeling in representation learning, and diffusion-based generative models with masking. 

\subsection{User Behavior Modeling for Mobile AI Applications}
Modeling user behavior is a long-standing problem in mobile AI, with app ranging from traffic forecasting and app usage prediction to proactive service delivery. 
Early work focused on \emph{aggregate traffic modeling} to predict total demand across base stations or geographic regions~\cite{xu2017real, awad2015support}.
Recent approaches replace statistical models with spatio-temporal graph neural networks to capture structured dependencies across regions and time~\cite{yu2017spatio, wu2019graph, hui2023large, zhou2024graph, liu2024spatial}. 
Infrastructure-level methods ignore individual heterogeneity by treating users as interchangeable and relying only on population signals. Personalized app prediction models next-app usage as a sequential task from past interactions~\cite{zhang2021dual, shi2020spatial}, with LSTMs~\cite{azari2019cellular} and graph-based approaches capturing temporal or app-to-app dependencies—but assuming behavior is fully history-determined, requiring dense logs and limiting cold-start and exploratory recommendations. Context-aware models add time, location, activity, and semantics~\cite{wang2021app2vec}, typically via handcrafted or feature-level fusion~\cite{huang2025appgen, duan2025attrappgen}, yet do not define which behaviors are predictable per user. Even attention-based spatio-temporal models~\cite{li2021spatiotemporal} lack individual-level alignment of spatial, temporal, and semantic cues, often favoring aggregate patterns~\cite{hui2023large, gong2025sttf}. Although large language models infer high-level intent from sparse signals~\cite{meng2025tuning, zeng2022glm}, current LLM-based methods use semantics only as auxiliary features or standalone predictors~\cite{meng2025tuning, zeng2022glm}, without mapping intent to an explicit spatio-temporal prediction structure.
As a result, semantics remain decoupled from the generative process and do not guide where or when prediction occurs, limiting effectiveness in cold-start and exploratory recommendation. 
This gap highlights the absence of a principled abstraction that connects user semantics to personalized inference structure in mobile AI.

\subsection{Masking Strategies in Representation Learning}
Masking strategy is central to self-supervised representation learning as it explicitly controls what is observed and what must be inferred.
By reconstructing partially observed inputs, models internalize structural priors governing token dependencies. This principle originated as masked language modeling in NLP~\cite{devlin2019bert} and was later extended to vision via masked autoencoders like MAE and BeiT~\cite{he2022masked, bao2021beit}, showing that the masking pattern fundamentally shapes the inductive bias of learned representations.
Masked modeling has been extended to spatio-temporal domains like video and multivariate time series~\cite{tong2022videomae, li2023unmasked}, capturing temporal continuity, periodicity, and cross-modal dependencies. Since a mask separates evidence from what must be inferred, masking strategies encode different assumptions about predictable dependencies. Random masking removes tokens independently to encourage global context use~\cite{he2022masked}, while block or tube masking removes contiguous regions to support local spatio-temporal inference~\cite{tong2022videomae}. Temporal masking enforces forecasting by hiding future segments, whereas autoregressive transformers achieve causality through attention constraints instead of input masking~\cite{devlin2019bert}.
Existing masking strategies are static and non-adaptive, applying the same pattern uniformly across samples under the assumption that inferable structure is identical for all. Masking acts mainly as a training heuristic or architectural prior rather than a mechanism for user-level spatio-temporal prediction support, limiting its use in personalized behavior modeling where relevance and predictability depend on individual semantics and task goals.

\subsection{Diffusion Models with Masking} 
Diffusion models excel at conditional synthesis under uncertainty by iteratively denoising to approximate complex distributions~\cite{ho2020denoising}. Unlike deterministic or autoregressive models that predict a single outcome, they capture multimodal futures and are well suited for behavior modeling where user actions remain stochastic and context-dependent despite similar histories.
Recent work combines masking with diffusion models for conditional completion from partial inputs~\cite{saharia2022palette, gungor2023adaptive}. Masking separates observed evidence from regions to generate, and the diffusion process operates over the full sequence conditioned on visible data. Observed entries are preserved at each denoising step to maintain context consistency while enabling flexible generation in masked areas, allowing diffusion models to serve as a general backbone for spatio-temporal content reconstruction across domains.
Existing diffusion-based completion methods assume a predefined static mask applied uniformly across samples, independent of user semantics or task goals. As a result, while they specify how to generate missing content, they rely on external masks to decide what to generate, limiting their ability to support personalized inference where spatio-temporal relevance depends on individual behavior and intent.

\section{Framework Overview}\label{sec:overview}
\begin{figure}[t]
\centering
\includegraphics[width=0.47\textwidth]{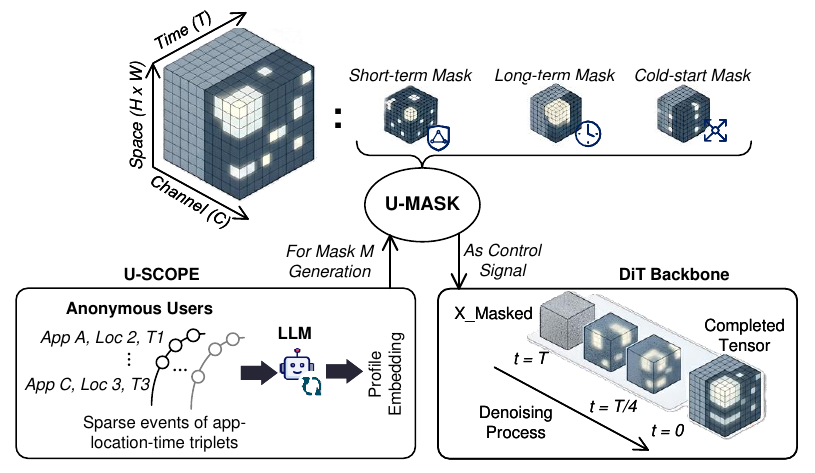}
\caption{Unified framework for personalized mobile AI behavior modeling.}
\label{unified}
\end{figure}

In this section, we introduce our unified framework for personalized mobile AI usage behavior modeling as illustrated in Fig.~\ref{unified}.

\subsection{Problem Setup and Behavioral Tensor}
\label{subsec:problem}

We consider personalized mobile AI apps operating under dynamic, sparse, and heterogeneous usage conditions. 
Unlike conventional prediction problems with fixed targets and fixed conditioning contexts, such app require the system to adapt its inference context jointly to the user and the task. 
Depending on the objective, inference may emphasize recent activity, long-term behavioral regularities, or robustness under extreme data sparsity in cold-start scenarios. 
As discussed in Section~\ref{sec:introduction}, these objectives impose conflicting requirements on how behavioral evidence should be selected and used. 
Our goal is therefore to develop a unified modeling framework that can explicitly adapt the inference context to both user characteristics and task demands.
Here, we formulate personalized mobile AI inference as a conditional completion problem. 
Given partial observations of a user's spatio-temporal behavior tensor $\mathcal{X}_u$, the objective is to infer its unobserved regions under task-specific requirements. Throughout the paper, we use the term \emph{inference} to denote this conditional completion process, rather than prediction of a fixed target variable.

For each user $u$, mobile behavior is represented as a partially observed spatio-temporal tensor
\begin{equation}
\mathcal{X}_u \in \mathbb{R}^{C \times T \times H \times W},
\end{equation}
where $C$ denotes behavior channels, $T$ indexes discrete time slots, and $H \times W$ defines a spatial tessellation.
Each entry corresponds to a behavioral signal at a specific time and location.
In real-world settings, observations are limited and spread unevenly because activity is intermittent and logging is restricted.

\subsection{Mask-Guided Personalized Inference}
\label{subsec:framework}
We realize personalized mobile AI inference through a mask-guided conditional completion framework composed of three tightly coupled components: a user- and task-specific masking mechanism (i.e., U-MASK), a semantic user representation module (i.e., U-SCOPE), and a shared conditional generative backbone based on DiT.

At the core of the framework is U-MASK, which defines inference by explicitly specifying which spatio-temporal regions of the behavior tensor are treated as observed evidence.
For each user $u$ and inference condition $\tau$, U-MASK produces a binary mask as
\begin{equation}
M_{u,\tau} \in \{0,1\}^{T \times H \times W},
\end{equation}
where $M_{u,\tau}(t,h,w)=1$ indicates that all $C$ behavior channels at location $(t,h,w)$ are revealed as evidence, and $M_{u,\tau}(t,h,w)=0$ indicates that the corresponding signals are unobserved and must be inferred.
Given $M_{u,\tau}$, inference is formulated as conditional completion as
\begin{equation}
p\!\left(\mathcal{X}_u \mid M_{u,\tau} \odot \mathcal{X}_u\right),
\end{equation}
where $\odot$ denotes element-wise masking with broadcast over the channel dimension.
Different inference objectives correspond to different mask structures, allowing short-term adaptation, long-horizon forecasting, and cold-start inference to be expressed within a single formulation.

To construct meaningful user-adaptive masks under sparse observations, U-MASK is conditioned on a compact semantic user representation as
\begin{equation}
\mathbf{h}_u \in \mathbb{R}^{d},
\end{equation}
produced by U-SCOPE.
U-SCOPE maps raw app-location interaction histories into a fixed-dimensional embedding that captures stable behavioral regularities and latent user intent.
This embedding is task-agnostic and serves as the sole semantic interface for personalization, modulating how U-MASK allocates evidence across time and space without introducing task-specific prediction heads.

Given $(M_{u,\tau}, \mathbf{h}_u)$, conditional completion is performed by a shared generative backbone based on a DiT architecture.
The DiT backbone models spatio-temporal dependencies in $\mathcal{X}_u$ and reconstructs unobserved regions while preserving the masked evidence.
Crucially, the backbone itself is user-agnostic and task-agnostic in its architecture.
All task differentiation and personalization enter exclusively through U-MASK and U-SCOPE, which cleanly separates task specification, user adaptation, and spatio-temporal dependency modeling.

The detailed designs of U-MASK, U-SCOPE, and the DiT backbone are presented in Sections~\ref{sec:umask}, \ref{sec:uscope}, and \ref{sec:diffusion}, respectively.

\section{U-MASK: User- and Task-Specific Spatio-Temporal Masking}
\label{sec:umask}
This section presents U-MASK, a user- and task-specific spatio-temporal masking framework. It defines how a user-level representation is translated into task-dependent decisions about which spatio-temporal regions should be revealed as evidence to support conditional completion.

\subsection{Hierarchical User Behavioral Representation}
\label{subsec:hierarchical}
U-MASK requires an intermediate representation that summarizes user behavioral regularities at multiple temporal scales and can be flexibly modulated by different inference objectives.
To this end, we construct a hierarchical behavioral representation from the spatio-temporal behavior tensor $\mathcal{X}_b$, while allowing task- and user-level semantics to modulate this representation.

Specifically, we first encode the observed behavior tensor $\mathcal{X}_b$ into a compact user-level embedding $\mathbf{h}_b \in \mathbb{R}^{d}$  using a shared behavioral encoder.
This encoder aggregates spatio-temporal patterns across time and location and produces a fixed-dimensional representation that captures the overall behavioral structure of user $u$.

U-MASK converts the semantic user embedding $\mathbf{h}_b \in \mathbb{R}^{d}$ into task-relevant latent factors that characterize behavioral regularities at different temporal scales.
This process is implemented by a trainable Transformer-based hierarchical encoder 
$\mathcal{T}_{\text{hier}} : \mathbb{R}^{d} \rightarrow \mathbb{R}^{d} \times \mathbb{R}^{d} \times \mathbb{R}^{d}$, which maps $\mathbf{h}_b$ to three latent vectors as
\begin{equation}
\label{eq:disentangle}
 \{\mathbf{F}^{\text{short}}_b,\ \mathbf{F}^{\text{long}}_b,\ \mathbf{F}^{\text{pat}}_b\}
= \mathcal{T}_{\text{hier}}(\mathbf{h}_b),
\end{equation}
corresponding to short-term activity fluctuations, long-term periodic trends, and pattern-level behavioral features, respectively. For each inference task $\tau$, these factors are combined to form a task-specific latent representation
\begin{equation}
\label{eq:task_proj}
\mathbf{F}_{b,\tau} = P_\tau\big[\mathbf{F}^{\text{short}}_b;\mathbf{F}^{\text{long}}_b;\mathbf{F}^{\text{pat}}_b\big] \in \mathbb{R}^{d_\tau},
\end{equation}
where $P_\tau$ is a learnable linear projection and $;$ denotes concatenation. 
This representation encodes how different temporal aspects of user behavior are emphasized under task $\tau$.
To estimate the relative importance of individual latent dimensions under each task, we introduce a task-dependent sensitivity predictor inspired by the diagonal of the Fisher information matrix. 
Specifically, for a given user $u$, we employ a lightweight sensitivity predictor $\mathcal{G}\tau(\cdot)$ to estimate the feature-wise importance.
We approximate the diagonal of the Fisher information matrix using the squared magnitude of this predictor's output as
\begin{equation}
\label{eq:fisher_approx}
\operatorname{diag}(I_{b,\tau}) \approx
\mathcal{G}_\tau(\mathbf{F}_{b,\tau})
\odot
\mathcal{G}_\tau(\mathbf{F}_{b,\tau})
+
\varepsilon \mathbf{1},
\end{equation}
where $\odot$ denotes the Hadamard product and $\varepsilon$ is a numerical stabilizer.
Theoretically, $\mathcal{G}\tau$ acts as a learnable gradient approximation of the objective function.
Since Fisher information reflects the squared gradient, our formulation effectively estimates the loss sensitivity to input perturbations.
High values in $\operatorname{diag}(I_{b,\tau})$ signal informative features, where even minor variations significantly impact predictions.
Crucially, unlike population-averaged standard methods, our approach yields instance-specific estimates to capture unique user patterns.
Finally, we define the latent importance profile as $\lambda_{b,\tau} = \sqrt{\operatorname{diag}(I_{b,\tau})}$ and use it as a weighting signal to guide the next step of generating a task specific mask.

\subsection{Task-Specific Feature Refinement}
\label{subsec:adaptation}

To account for task-specific variations in stable behavioral structure without re-learning user representations we introduce a lightweight refinement mechanism. 
This process operates exclusively on the pattern-level feature $\mathbf{F}^{\text{pat}}_{b}$ as it encodes long-term behavioral regularities that directly influence the spatial affinity modeling in U-MASK.
Formally we initialize the refinement state as $\mathbf{Z}_{b,0}=\mathbf{F}^{\text{pat}}_{b}$ and perform $K$ unrolled refinement steps defined as
\begin{equation}
\mathbf{Z}_{b,k+1}
=
\mathbf{Z}_{b,k}
+
\eta\,\Delta_{\tau}(\mathbf{Z}_{b,k}),
\quad
k=0,\ldots,K-1,
\label{eq:refinement_umask}
\end{equation}
where $\Delta_{\tau}(\cdot)$ is a small task-specific residual network parameterized independently for each task condition $\tau$ and $\eta>0$ is a fixed step size controlling the update magnitude.
The final refined feature $\mathbf{Z}_{b}=\mathbf{Z}_{b,K}$ replaces the static $\mathbf{F}^{\text{pat}}_{b}$ in the subsequent spatial affinity estimation (Eq.~\ref{spatial affinity}). 
This mechanism allows the model to dynamically adjust the user's historical profile to better align with the specific requirements of the current inference task.

\subsection{Task-specific Spatio-Temporal Mask Generation}
\label{subsec:mask_gen}
For each user $u_b$ and task condition $\tau$, U-MASK generates the mask $M_{b,\tau}$ through three coupled components: (i) an \emph{observation ratio} that allocates the evidence budget, (ii) a \emph{spatio-temporal sampling distribution} that ranks candidate coordinates, and (iii) a \emph{ratio-constrained sampler} that produces a binary mask satisfying the budget.

\subsubsection{Observation ratio}
We first determine a target observation ratio $\tilde{\rho}_{b,\tau}\in(0,1)$, which controls the number of spatio-temporal coordinates revealed as evidence. 
For each $u_b\in\mathcal{B}$, we stack pattern-level latent vector $\mathbf{Z}_{b}\in\mathbb{R}^{d}$ as 
\begin{equation}
\mathbf{Z}_{\mathcal{B}}=\big[\mathbf{Z}_{b_1};\ldots;\mathbf{Z}_{b_n}\big]\in\mathbb{R}^{B\times d}.
\end{equation}
U-MASK uses $\mathbf{Z}_{\mathcal{B}}$ to estimate how reliably the $b$-th user $u_b$ can borrow statistical strength from other users, which in turn controls how much evidence should be revealed for that user.

\textbf{Batch-wise similarity and reliability.}
We first map $\mathbf{Z}_{\mathcal{B}}$ to $K$ soft behavior groups using a learnable assignment network $\Phi(\cdot)$. 
The soft assignment matrix $\mathbf{A} \in \mathbb{R}^{B\times K}$ is computed as
\begin{equation}
\mathbf{A}=\mathrm{softmax}\!\left(\frac{\Phi(\mathbf{Z}_{\mathcal{B}})}{T_{\mathrm{tmp}}}\right),
\label{eq:assignment_umask_polish}
\end{equation}
where the temperature scalar $T_{\mathrm{tmp}}$ controls the sharpness of the group distribution. 
Each row $\mathbf{A}_{b}$ represents a simplex vector encoding the group membership of user $u_b$. 
We then introduce a similarity matrix for each batch $\mathbf{S} \in \mathbb{R}^{B\times B}$ defined as follows
\begin{equation}
\mathbf{S}=\alpha\,\mathbf{A}\mathbf{A}^{\top}+(1-\alpha)\,\mathbf{U}\mathbf{U}^{\top},
\label{eq:similarity_umask_polish}
\end{equation}
where $\mathbf{U}$ stacks the $\ell_{2}$-normalized feature vectors and $\alpha$ is a learnable mixing coefficient within the unit interval. 
The first term measures similarity through shared group membership while the second term captures continuous alignment in the representation space. 
For each user $u_b$ we identify the strongest neighbor similarity as $s_b=\max_{j\neq b}\mathbf{S}_{b,j}$. 
A larger $s_b$ means the user has similar peers in the batch and can depend on shared structures instead of relying only on direct evidence.

\textbf{User-specific scaling factor.}
To translate these statistics into an evidence budget we construct a context vector $\mathbf{e}_b \in \mathbb{R}^3$ containing the neighbor similarity $s_b$, the feature norm $\|\mathbf{Z}_{b}\|_2$ and the assignment entropy $H(\mathbf{A}_{b})=-\sum_{k=1}^{K}\mathbf{A}_{b,k}\log \mathbf{A}_{b,k}$.
We then compute a user-specific scaling factor $\gamma_b$ using a learnable projection defined by
\begin{equation}
\label{scaling_factor_umask}
\gamma_b=\sigma(\mathbf{w}^{\top}\mathbf{e}_b+c),
\end{equation}
where $\mathbf{w}$ and $c$ are learnable parameters and $\sigma(\cdot)$ is the sigmoid function. 
This factor $\gamma_b$ dynamically modulates the observation budget. 
Through end-to-end training the network learns to map low neighbor similarity or high assignment uncertainty to a larger $\gamma_b$. 
This mechanism effectively allocates a higher ratio of observed evidence to isolated users who lack strong statistical support from the batch while applying more aggressive masking for users with redundant information.

\textbf{Task-specific base ratio.}
Different inference tasks naturally require varying amounts of observed information. 
Short-term prediction and long-term prediction impose distinct data demands. 
Calculating gradients during the inference stage is computationally impractical. 
Therefore we introduce a learnable task-specific base ratio $\rho_{\tau}$. 
This parameter takes a value between zero and one. 
We optimize it jointly with the network weights. 
This approach allows the model to automatically learn the ideal amount of visible data that balances task difficulty and reconstruction accuracy for each condition $\tau$.

\textbf{observation ratio.}
We synthesize the global task demand and the local user reliability to determine the final target ratio $\tilde{\rho}_{b,\tau}$. 
Given the task base ratio $\rho_{\tau}$ and the user scaling factor $\gamma_b$ derived above we define
\begin{equation}
\tilde{\rho}_{b,\tau} = \rho_{\tau} \cdot \gamma_b.
\label{eq:ratio_final_polish}
\end{equation}
Thus the model generates a denser mask for easy tasks or well-supported users while preserving more observed coordinates for difficult tasks or isolated users.

\subsubsection{Spatio-Temporal Sampling Distribution}
\label{subsubsec:sampling_dist}

Given the target observation ratio $\tilde{\rho}_{b,\tau}$ derived above U-MASK defines a sampling distribution to rank candidate coordinates based on their utility. 
We assign each spatio-temporal coordinate $(t,h,w)$ a relevance score $p_{t,h,w}^{(b,\tau)}$ as
\begin{equation}
p_{t,h,w}^{(b,\tau)} = \beta_{\tau}\,\phi_{t}^{(\tau)} + \big(1-\beta_{\tau}\big)\,\psi_{h,w}^{(b)},
\label{eq:saliency_umask_polish}
\end{equation}
where $\phi_{t}^{(\tau)}$ captures task-specific temporal dynamics while $\psi_{h,w}^{(b)}$ encodes user-specific spatial affinity.

\textbf{Temporal importance.}
Certain tasks rely heavily on specific recent time steps. 
We define $\phi^{(\tau)} \in \mathbb{R}^T$ as a learnable temporal profile vector optimized for task $\tau$. 
The element $\phi_{t}^{(\tau)}$ represents the learned global importance of the $t$-th time step which allows the model to automatically identify temporal segments that are most informative for the current inference goal.

\textbf{Feature-weighted spatial affinity.}
We incorporate the latent importance profile $\lambda_{b,\tau}$ derived in Section~\ref{sec:umask} to guide spatial exploration.
The profile $\lambda_{b,\tau}$ captures the task-specific sensitivity of each feature dimension. 
Leveraging these weights we compute a feature-weighted compatibility between the user pattern vector $\mathbf{Z}_{b}$ and the location embedding $\mathbf{E}_{\mathrm{pos}}(h,w)$. 
Here, the location embedding \(E_{\text{pos}}(h, w)\) represents the clustered label of the user’s location, derived from the surrounding Points of Interest (POIs), where the user’s location information is inferred via the serving base station to which the user is connected.
The spatial affinity is defined as the rectified weighted cosine similarity:
\begin{equation}
\psi_{h,w}^{(b)}=
\mathrm{ReLU}\!\left(
\frac{(\mathbf{Z}_{b} \odot \lambda_{b,\tau})^{\top}\mathbf{E}_{\mathrm{pos}}(h,w)}
{\|\mathbf{Z}_{b} \odot \lambda_{b,\tau}\|_{2}\,\|\mathbf{E}_{\mathrm{pos}}(h,w)\|_{2}}
\right).
\label{spatial affinity}
\end{equation}
This mechanism ensures priority for locations that match user behavior in aspects that are crucial for task $\tau$.
Such selectivity lets weekend prediction tasks focus more on leisure places instead of work sites.

\textbf{Task-dependent mixing.}
The mixing coefficient $\beta_{\tau} \in (0,1)$ is predicted from the task embedding $\mathbf{e}_{\tau}$ using $\beta_{\tau}=\sigma(\mathbf{v}^{\top}\mathbf{e}_{\tau} + c_{\tau})$ so that it balances the temporal signal and the spatial signal.

\subsubsection{Ratio-Constrained Mask Sampling}
\label{subsubsec:mask_sampling}
We convert continuous relevance scores into a discrete weighted mask using the target observation ratio $\tilde{\rho}_{b,\tau}$ derived in Eq.~\ref{eq:ratio_final_polish}.
Let $N=THW$ be the total number of coordinates. 
We strictly enforce the sparsity constraint by calculating the evidence budget as $k_b = \lfloor \tilde{\rho}_{b,\tau} N \rfloor$. 
We employ the Gumbel-Top trick $k$ to generate a binary selection map $I_{b,\tau}$ that identifies the top coordinates $k_b$ most relevant:
\begin{equation}
\label{eq:one_hot}
I_{b,\tau} = \mathrm{OneHot}\!\left(\mathrm{GumbelTopK}\!\left(\log p^{(b,\tau)} + G, \ k_b\right)\right).
\end{equation}
We construct the importance-weighted mask $\mathcal{M}_{b,\tau}$ by modulating the selected indices with their relevance scores
\begin{equation}
\label{eq:soft mask}
\mathcal{M'}_{b,\tau} = I_{b,\tau} \odot p^{(b,\tau)}.
\end{equation}
This formulation effectively integrates the quantity constraint from $\tilde{\rho}_{b,\tau}$ and the quality guidance from $\lambda_{b,\tau}$.
Specifically, $\lambda_{b,\tau}$ modulates the spatial affinity component within $p^{(b,\tau)}$ thereby suppressing locations that do not align with task-critical features.
The resulting mask acts as a soft denoising filter that preserves only the most informative spatio-temporal fragments for the subsequent encoder.

\subsubsection{Cold-Start Adjustment}
\label{subsubsec:cold_start}

For the cold-start condition $\tau=\mathrm{cold}$ the historical evidence is sparse which makes the estimated spatial affinity $\psi_{h,w}^{(b)}$ unreliable. 
We enforce exploration to prevent overfitting to noisy history by flattening the sampling distribution. 
We define the adjusted distribution as a mixture with a uniform prior given by
\begin{equation}
\label{eq:cold_start_adj}
\tilde{p}_{t,h,w}^{(b,\mathrm{cold})}
=
(1-\epsilon)\,p_{t,h,w}^{(b,\mathrm{cold})}
+
\epsilon\,\frac{1}{THW},
\end{equation}
where $\epsilon \in [0,1]$ is a hyperparameter controlling exploration strength. 
This mechanism maximizes the entropy of the mask and encourages the model to attend to a broader range of contextual cues rather than over-trusting specific regions.

\subsection{End-to-End Training Objective}
\label{subsec:training_umask}

U-MASK is trained jointly with the conditional generative backbone described in Section~\ref{sec:diffusion}.
Let $\mathcal{X}_{b,t}$ denote the noisy behavior tensor of user $b$ at diffusion step $t$ and $\epsilon_{\theta}(\cdot)$ denote the denoised output of the DiT backbone.
We define the primary reconstruction objective as
\begin{equation}
\mathcal{L}_{\mathrm{rec}}
=
\mathbb{E}_{b,\tau,t}
\left[
\left\|
\mathcal{X}_{b,0}
-
\epsilon_{\theta}\!\left(
\mathcal{M'}_{b,\tau} \odot \mathcal{X}_{b,t},\,
\mathbf{F}_{b,\tau},\,
t,\,
\tau
\right)
\right\|_{2}^{2}
\right].
\label{eq:loss_rec_umask}
\end{equation}
Here the weighted mask $\mathcal{M}_{b,\tau}$ filters the input noise based on importance scores while $\mathbf{F}_{b,\tau}$ provides the task-adapted user context.
To encourage coherence between short-term fluctuations and long-term trends we add a representation consistency regularizer:
\begin{equation}
\mathcal{L}
=
\mathcal{L}_{\mathrm{rec}}
+
\lambda_{\mathrm{con}}\,
\mathcal{L}_{\mathrm{InfoNCE}}
\!\left(
\mathbf{F}^{\text{short}}_{b},
\mathbf{F}^{\text{long}}_{b}
\right),
\label{eq:loss_umask}
\end{equation}
where $\lambda_{\mathrm{con}}>0$ is a weighting coefficient.
The term $\mathcal{L}_{\mathrm{InfoNCE}}$ is computed over users within a mini-batch to explicitly align the local short-term activities with global long-term preferences.
Since the sampling operation in U-MASK relies on the continuous relaxation of the Top-$k$ operator gradients from $\mathcal{L}_{\mathrm{rec}}$ propagate through the mask generator to optimize the observation ratio predictor and the temporal importance profile end-to-end.

\section{U-SCOPE: Semantic User Representation for Personalized Mobile Behavior Modeling}
\label{sec:uscope}
\begin{figure}[t]
\centering
\includegraphics[width=0.47\textwidth]{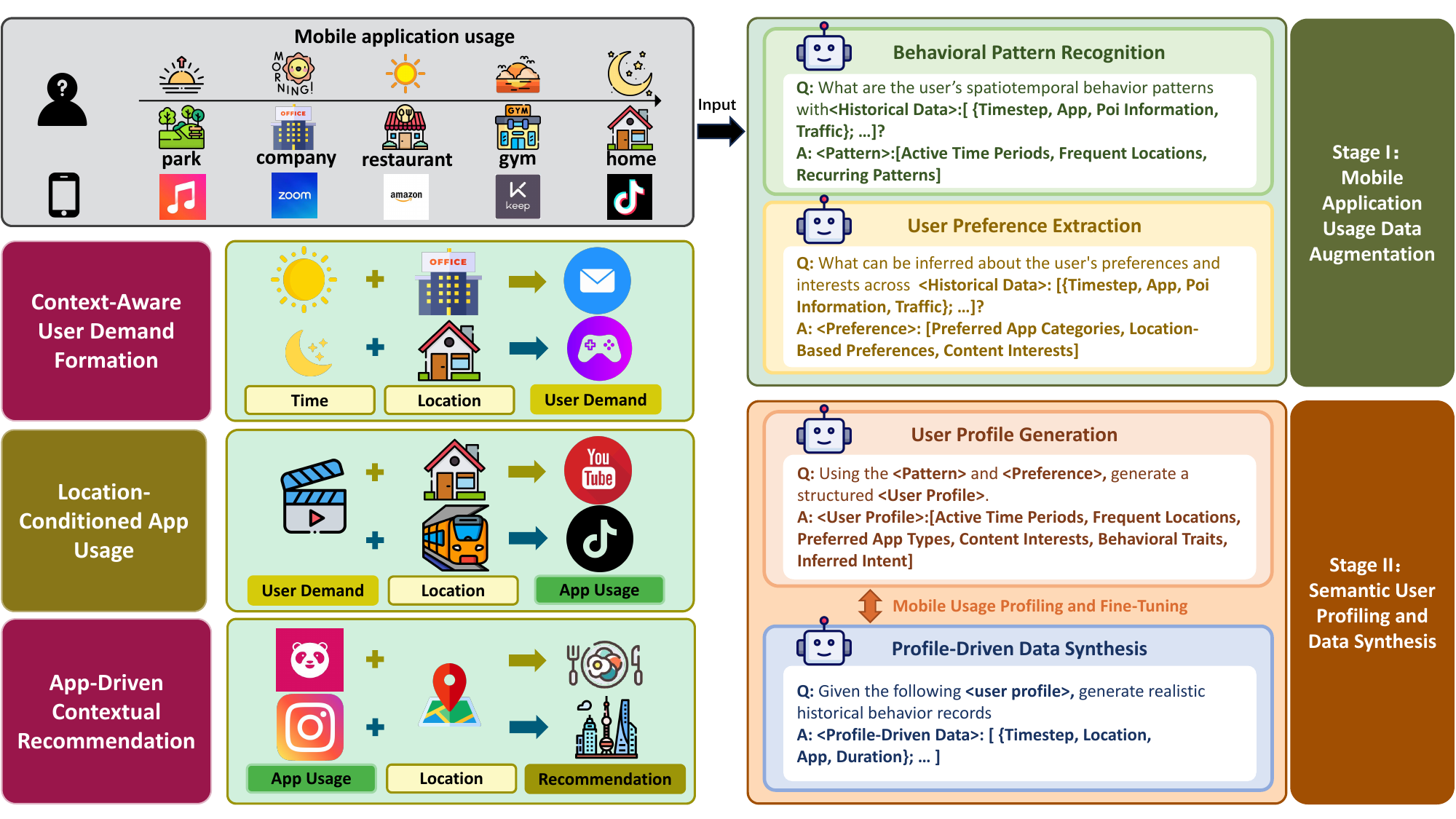}
\caption{Overview of the U-SCOPE framework. Given sparse mobile application usage data, U-SCOPE infers semantic user representations that encode application preferences, location tendencies, and temporal regularities, enabling downstream task-oriented inference under data scarcity.}
\label{fig:framework}
\end{figure}

U-MASK provides a principled mechanism for translating user-level representations into task-adaptive spatio-temporal masks. However, its effectiveness is affected by the quality of the user representation used to condition mask construction, i.e., $\mathbf{h}_u$. In mobile settings, raw app-location interaction data are highly sparse, fragmented, and heterogeneous, making it difficult to directly infer stable behavioral semantics from $\mathcal{X}_u$ alone.
This motivates the design of U-SCOPE as the perception component of our framework. 
It is a semantic user profiling component that infers compact and task-agnostic user representations from sparse mobile telemetry as shown in Fig.~\ref{fig:framework}.
The inferred embedding is used exclusively to condition U-MASK and does not define any task-specific prediction head.
As a result, all personalization and task differentiation in our framework are mediated through spatio-temporal masking rather than specialized output modules.

\subsection{LLM-Grounded Semantic User Representation}
Inferring user intent from sparse behavioral observations is challenging due to limited historical coverage, missing context, and the absence of explicit supervision. U-SCOPE addresses this challenge by treating LLMs as amortized semantic inference engines that map low-level mobile telemetry into structured representations of latent user intent.

We represent the observed behavior of user $u$ as a spatio-temporal event sequence
\begin{equation}
H_u = \{(a_t, \ell_t, \tau_t)\}_{t=1}^{T},
\end{equation}
where $a_t \in \mathcal{A}$ denotes the accessed app, $\ell_t \in \mathcal{L}$ denotes the visited location, and $\tau_t$ denotes the timestamp.
We conceptually introduce a latent semantic variable $\mathcal{P}_u$ that captures stable user intent and long-term behavioral preference.
Conditioned on $H_u$, an energy-based perspective~\cite{LeCun2006ATO} admits an unnormalized Gibbs form as
\begin{equation}
p(\mathcal{P}_u \mid H_u) \propto \exp\!\big( -\mathcal{E}(H_u, \mathcal{P}_u) \big),
\label{eq:profile_posterior}
\end{equation}
where $\mathcal{E}(\cdot)$ measures semantic compatibility between observed behavior and a hypothesized intent profile.
This formulation provides an interpretive probabilistic view of semantic consistency rather than an explicitly optimized objective.
In practice, U-SCOPE approximates this posterior through amortized inference~\cite{Lopez2020AUTOENCODINGVB,Hu2023AmortizingII}.
We construct a structured prompt $x_u$ that summarizes user behavior using temporal activity statistics $\pi^{\text{time}}_u$, app-location co-occurrence patterns $C_u \in \mathbb{R}^{|\mathcal{L}| \times |\mathcal{A}|}$, and session regularity indicators $\sigma_u$.
Conditioned on $x_u$, a pretrained LLM produces a natural-language semantic profile as
\begin{equation}
y_u \sim \pi_{\theta}(y \mid x_u),
\end{equation}
where $\pi_{\theta}$ denotes the LLM policy. The generated text $y_u$ is not treated as a prediction target. Instead, it serves as an intermediate semantic representation that externalizes latent intent in a structured form suitable for downstream processing. 
To obtain a compact and differentiable user representation, we encode $y_u$ using a lightweight semantic encoder as
\begin{equation}
\mathbf{h}_u = \text{MLP}\big( \texttt{[CLS]}_{\text{enc}}(y_u) \big) \in \mathbb{R}^{d},
\label{eq:user_embedding}
\end{equation}
where $\texttt{[CLS]}_{\text{enc}}(\cdot)$ denotes the [CLS] embedding extracted from a frozen Sentence-BERT encoder~\cite{reimers2019sentence}.
The resulting vector $\mathbf{h}_u$ is a task-agnostic semantic summary of user behavior and serves as the sole interface between U-SCOPE and U-MASK.

\subsection{Semantic-Augmented Behavioral Synthesis}
Under extreme data sparsity, semantic inference may become unstable due to insufficient coverage of user behavior.
To mitigate this issue during training, U-SCOPE optionally augments observed user traces with synthetic behavior that is semantically consistent with the inferred profile.
This mechanism functions as a regularizer and is not required during inference.
Formally, we define the synthesis process as a conditional distribution
\begin{equation}
p(H^{\mathrm{aug}}_u \mid \mathcal{P}_u) = \prod_{t=1}^{T'} p(a_t, \ell_t \mid \mathcal{P}_u, \tau_t),
\label{eq:synthesis}
\end{equation}
implemented by fine-tuning an LLM on real app-location sequences using next-token prediction, followed by constrained decoding~\cite{Geng2023GrammarConstrainedDF,Roy2024FLAPFP}.
During generation, app-location pairs with zero empirical co-occurrence in $H_u$ are disallowed unless exploratory intent is implied by the semantic profile.

To preserve structural consistency, we enforce an app-location correlation constraint as
\begin{equation}
\mathbb{E}_{H^{\mathrm{aug}}_u} \big[ \mathrm{Corr}(a, \ell) \big] \approx \mathrm{Corr}_{\mathrm{real}}(a, \ell),
\label{eq:correlation_preserve}
\end{equation}
where $\mathrm{Corr}(\cdot,\cdot)$ denotes empirical correlation.
The augmented set
$ 
\tilde{H}_u = H_u \cup \{ H^{\mathrm{aug}}_u \}_{i=1}^{K}
 $
is used only during training to stabilize semantic representation learning.

\subsection{Preference Alignment and Downstream Use}
Due to privacy and edge efficiency limits, large models are not suitable so U-SCOPE uses a small model for preference alignment.
To further improve robustness of semantic inference, U-SCOPE optionally incorporates a preference alignment step based on Direct Preference Optimization~\cite{rafailov2023direct,Sun2024DirectPO}.
A preference triplet $(x, y_w, y_l)$ consists of a contextual query $x$ derived from user history, a preferred response $y_w$ that is semantically consistent with $H_u$, and a dispreferred response $y_l$ generated by a heuristic or mismatched baseline.
The optimization objective is
\begin{equation}
\mathcal{L}_{\mathrm{DPO}} \! = \!
-\mathbb{E}\!\left[
\log \sigma\!\left( \!
\beta \! \left[ \log \frac{\pi_\theta(y_w|x)}{\pi_{\mathrm{ref}}(y_w|x)}
\!-\!
\log \frac{\pi_\theta(y_l|x)}{\pi_{\mathrm{ref}}(y_l|x)} \right]
\right)
\right]\!,
\label{eq:dpo}
\end{equation}
where $\pi_{\mathrm{ref}}$ is a frozen reference model and $\beta > 0$ controls the optimization sharpness.

Preference alignment encourages semantically consistent interpretations of sparse behavioral evidence while avoiding overfitting to limited histories.
Importantly, this step does not introduce additional prediction tasks or modify the generative backbone.
The output of U-SCOPE remains the compact embedding $\mathbf{h}_u$ defined in~\eqref{eq:user_embedding}, which serves as the sole semantic conditioning signal for U-MASK.
All downstream tasks rely solely on task oriented masking and generative completion. This includes short term prediction, long term forecasting and cold start recommendation.

\section{DiT Backbone: Mask-Guided Generative Modeling with Diffusion}
\label{sec:diffusion}
This section presents the shared DiT backbone acts as a general-purpose completion engine that reconstructs unobserved spatio-temporal mobile behavior while preserving observed evidence.


\subsection{Problem Setup and Conditioning Interface}
For each user $u$, mobile behavior is represented as a spatio-temporal tensor
$\mathcal{X}_u \in \mathbb{R}^{C \times T \times H \times W}$.
For a given inference condition $\tau$, U-MASK provides a binary mask $M_{u,\tau} \in \{0,1\}^{T \times H \times W}$, which is broadcast across the channel dimension.
Observed evidence is defined as
\begin{equation}
\mathcal{X}_u^{\mathrm{obs}} = M_{u,\tau} \odot \mathcal{X}_u,
\end{equation}
where $\odot$ denotes element-wise masking with implicit broadcast over $C$.

The DiT backbone models the conditional distribution
\begin{equation}
p\!\left(
\mathcal{X}_u \mid \mathcal{X}_u^{\mathrm{obs}}, M_{u,\tau}, \mathbf{h}_u, \tau
\right),
\end{equation}
and is used to infer missing spatio-temporal behavior given observed evidence and semantic context.

\subsection{Forward Diffusion and Conditional Denoising}
We adopt a standard variance-preserving diffusion process with horizon $T_d$.
Let $\{\beta_t\}_{t=1}^{T_d}$ denote a predefined noise schedule, and define $\alpha_t = 1 - \beta_t$ and $\bar{\alpha}_t = \prod_{s=1}^{t}\alpha_s$.
The forward process is defined as
\begin{equation}
q(\mathcal{X}_{u,t} \mid \mathcal{X}_{u,t-1})
=
\mathcal{N}\!\left(
\sqrt{\alpha_t}\,\mathcal{X}_{u,t-1},\,
(1-\alpha_t)\mathbf{I}
\right),
\end{equation}
which yields the closed-form expression
\begin{equation}
\mathcal{X}_{u,t}
=
\sqrt{\bar{\alpha}_t}\,\mathcal{X}_u
+
\sqrt{1-\bar{\alpha}_t}\,\boldsymbol{\epsilon},
\quad
\boldsymbol{\epsilon}\sim\mathcal{N}(\mathbf{0},\mathbf{I}).
\label{eq:forward_closed_form}
\end{equation}
We parameterize a denoiser $\epsilon_{\theta}$ that predicts the injected noise conditioned on observed evidence and semantic context.
The denoiser takes as input the noisy tensor $\mathcal{X}_{b,t}$ together with $(\mathcal{X}_b^{\mathrm{obs}}, M_{b,\tau}, \mathbf{h}_b, \tau, t)$.
Training minimizes the target reconstruction loss as
\begin{equation}
\mathcal{L}_{\mathrm{diff}}
=
\mathbb{E}_{b,\tau,t}
\left[
\left\|
\mathcal{X}_{b,0}
-
\epsilon_{\theta}\!\left(
\mathcal{M'}_{b,\tau} \odot \mathcal{X}_{b,t},\,
\mathbf{h}_{b,\tau},\,
t,\,
\tau
\right)
\right\|_{2}^{2}
\right].
\label{eq:diffusion_loss}
\end{equation}

Observed evidence is incorporated in two complementary ways.
First, $\mathcal{X}_b^{\mathrm{obs}}$ is provided explicitly as a conditioning input.
Second, evidence consistency is enforced during sampling through mask-guided clamping, described below.

\subsection{Diffusion Transformer Architecture}

The denoiser $\epsilon_{\theta}$ is implemented using a DiT that operates on spatio-temporal tokens.

\textbf{Spatio-temporal patching.}
We partition $\mathcal{X}_{u,t}$ into non-overlapping patches of size $(t_0,h_0,w_0)$, yielding a token sequence of length
$ 
L = (T/t_0)(H/h_0)(W/w_0).
$
Each patch is flattened and projected into a $D$-dimensional embedding, forming
$ 
\mathbf{E}_{u,t} \in \mathbb{R}^{L \times D}.
$
The same patching operation is applied to $\mathcal{X}_u^{\mathrm{obs}}$ to obtain aligned evidence tokens $\mathbf{E}_u^{\mathrm{obs}}$.

\textbf{Positional and temporal encoding.}
We add spatio-temporal positional embeddings and a diffusion timestep embedding as
\begin{equation}
\mathbf{Z}_{u,t}^{(0)} = \mathbf{E}_{u,t} + \mathbf{E}_{\mathrm{pos}} + \mathbf{E}_{\mathrm{time}}(t).
\end{equation}

\textbf{Semantic and task conditioning.}
The user embedding $\mathbf{h}_u$ and task identifier $\tau$ are projected to the model dimension, yielding embeddings $\mathbf{e}_u$ and $\mathbf{e}_{\tau}$.
We form a global conditioning vector as
\begin{equation}
\mathbf{c}_{u,\tau} = \mathbf{e}_u + \mathbf{e}_{\tau},
\end{equation}
which modulates Transformer blocks via adaptive layer normalization.
Observed evidence tokens $\mathbf{E}_u^{\mathrm{obs}}$ are fused through token-wise concatenation or cross-attention, providing explicit conditioning aligned with the mask.

\textbf{Transformer backbone.}
The DiT consists of $N$ Transformer blocks with multi-head self-attention and feed-forward layers.
The output tokens are projected back to the data space to produce $\hat{\boldsymbol{\epsilon}} \in \mathbb{R}^{C \times T \times H \times W}$.

\subsection{Mask-Guided Sampling}
\label{subsec:mask_guided_sampling}

At inference time, sampling proceeds from $\mathcal{X}_{u,T_d} \sim \mathcal{N}(\mathbf{0},\mathbf{I})$.
For $t=T_d,\ldots,1$, a reverse diffusion update produces a proposal $\tilde{\mathcal{X}}_{u,t-1}$.
We then enforce evidence consistency by clamping observed entries as
\begin{equation}
\mathcal{X}_{u,t-1}
=
(1-M_{u,\tau})\odot \tilde{\mathcal{X}}_{u,t-1}
+
M_{u,\tau}\odot \mathcal{X}_u,
\label{eq:evidence_clamp}
\end{equation}
where broadcasting across $C$ is implicit.
This operation ensures that the diffusion process only reconstructs masked regions and never alters observed evidence.
In cold-start settings, $M_{u,\tau}$ may reveal few or no entries, in which case generation is guided primarily by $(\mathbf{h}_u,\tau)$.

\begin{algorithm}[t]
\caption{U-MASK Joint Training Algorithm}
\label{alg:umask_training}
\renewcommand{\baselinestretch}{0.9}\selectfont
\begin{flushleft}
    \textbf{Input:}\\
    \quad $\bullet$ Training dataset\\
    \qquad - User Identifier $u_b$: Identifier for user-specific semantic retrieval\\
    \qquad - Behavior Tensor $\mathcal{X}_b$: Raw behavioral signals.\\
    \qquad - Task Type $\tau$: Identifier for inference objective\\
    \quad $\bullet$ Location Embeddings $\mathbf{E}_{\mathrm{pos}}$: Embedding matrix representing the spatial coordinates.\\
    \textbf{Learnable Parameters}: \\
    \quad $\bullet$ Encoder: Behavioral encoder \& Hierarchical projector \\
    \quad $\bullet$ Sensitivity: Fisher predictors $G_{\tau}, P_{\tau}$ \\
    \quad $\bullet$ Refinement: Residual network $\Delta_{\tau}$ \\
    \quad $\bullet$ Policy: Ratio predictor $\gamma_b$, Base ratio $\rho_{\tau}$, Temporal profile $\phi^{(\tau)}$, Mixing $\beta_{\tau}$\\
    \quad $\bullet$ Backbone: DiT Denoiser $\epsilon_{\theta}$ \\
    \textbf{Optimization Procedure:}
\end{flushleft}
\vspace{-0.2cm}
\begin{algorithmic}[1]
\State Initialize all parameters randomly
\While{not converged}
    \State Sample mini-batch $\{(\mathcal{X}_{b,0}, u_b, \tau)\}$ from $\mathcal{D}$
    \State Extract latent factors: $\{\mathbf{F}^{\text{short}}, \mathbf{F}^{\text{long}}, \mathbf{F}^{\text{pat}}\}$ according to (\ref{eq:disentangle})
    \State Compute task context $\mathbf{F}_{b,\tau}$ and sensitivity $\lambda_{b,\tau}$ according to (\ref{eq:task_proj}, \ref{eq:fisher_approx})
    \State Iterative residual refinement of pattern features to obtain ${Z}_b$ according to (\ref{eq:refinement_umask})
    \State Predict user scaling $\gamma_b$ and retrieve task base ratio $\rho_{\tau}$ and get final observation ratio $\tilde{\rho}_{b,\tau}$ according to (\ref{scaling_factor_umask},  \ref{eq:ratio_final_polish})
    \State Set budget: $k_b \leftarrow \lfloor \tilde{\rho}_{b,\tau} N \rfloor$
    \State Compute spatial affinity $\psi^{(b)}$ using $\mathbf{Z}_{b}, \lambda_{b,\tau}$ and $\mathbf{E}_{\mathrm{pos}}$ according to (\ref{spatial affinity})
    \State Retrieve temporal profile $\phi^{(\tau)}$ and mixing gate $\beta_{\tau}$
    \State Generate final relevance scores $p^{(b,\tau)} $ according to (\ref{eq:saliency_umask_polish})
    \If{$\tau = cold$}
        \State Adjust distribution $\tilde{p}_{t,h,w}^{(b,\mathrm{cold})}$ with uniform prior exploration according to (\ref{eq:cold_start_adj})
    \EndIf
    \State Sample Gumbel noise $G \sim \mathrm{Gumbel}(0,1)$
    \State Obtain binary mask $I_{b,\tau}$ and corresponding soft mask $\mathcal{M}_{b,\tau}$ according to (\ref{eq:one_hot}, \ref{eq:soft mask})
    \State \textbf{Noise Sampling and Target Prediction}
    \State \quad $\circ$ Sample time step $t$ and noise $\epsilon \sim \mathcal{N}(\mathbf{0}, \mathbf{I})$
    \State \quad $\circ$ $\mathcal{X}_{b,t} \leftarrow \sqrt{\bar{\alpha}_t}\mathcal{X}_{b,0} + \sqrt{1-\bar{\alpha}_t}\epsilon$ \Comment{Add Noise}
    \State \quad $\circ$ $\mathcal{X}^{\text{obs}}_{b} \leftarrow \mathcal{M}_{b,\tau} \odot \mathcal{X}_{b,t}$ \Comment{Apply Adaptive Mask}
    \State \quad $\circ$ Predict target: $\hat{\mathcal{X}} \leftarrow \epsilon_{\theta}(\mathcal{X}_{b,t}, \mathcal{X}^{\text{obs}}_{b}, \mathbf{F}_{b,\tau}, t,\tau)$
    
    \State Joint objective combining diffusion reconstruction loss with consistency regularization according to (\ref{eq:loss_rec_umask}, \ref{eq:loss_umask})
\EndWhile
\end{algorithmic}
\end{algorithm}

\subsection{Joint Optimization with U-MASK and U-SCOPE}

The DiT backbone is trained jointly with U-MASK and U-SCOPE.
U-MASK supplies task- and user-specific masks, while U-SCOPE supplies semantic user embeddings.
Gradients from the diffusion loss in Eq.~\eqref{eq:diffusion_loss} propagate through both modules, aligning mask construction and semantic representation learning with generative completion quality.
The training process and the way these components interact are described in Alg.~\ref{alg:umask_training}.
Despite this coupling, the DiT backbone itself remains shared and task-agnostic, serving as a unified completion engine across all inference regimes.

\section{Experimental Evaluation}
\label{sec:exp}
We conduct a comprehensive experimental evaluation of the proposed framework by considering the following Research Questions (RQs):

$\bullet$ \textbf{RQ1:} How does U-MASK enable robust and reliable multi-task learning?

$\bullet$ \textbf{RQ2:} How does U-MASK perform compared to conventional spatio-temporal masking?

$\bullet$ \textbf{RQ3:} How does U-SCOPE preserve core behavioral correlations in generated user data?

$\bullet$ \textbf{RQ4:} How does the effectiveness of user profiles generated through the U-SCOPE?

$\bullet$ \textbf{RQ5:} How does fine-tuning with synthetic user data from U-SCOPE improve their ability to address the cold-start problem?

$\bullet$ \textbf{RQ6:} How does the task-specific feature refinement module adapt spatio-temporal feature selection to the granularity of the task?

$\bullet$ \textbf{RQ7:} How does U-MASK generate personalized masking strategies from latent user behavioral patterns?

Collectively, these research questions evaluate the framework at three levels. RQ1–RQ2 examine the effectiveness of U-MASK, RQ3–RQ5 assess how U-SCOPE provides informative semantic signals to U-MASK, and RQ6–RQ7 analyze how the U-MASK–generated adaptive masks are exploited by the DiT backbone.

\subsection{RQ1: Multi-Task Learning with U-MASK}

\begin{figure*}[t]
\centering
\includegraphics[width=1.0\textwidth]{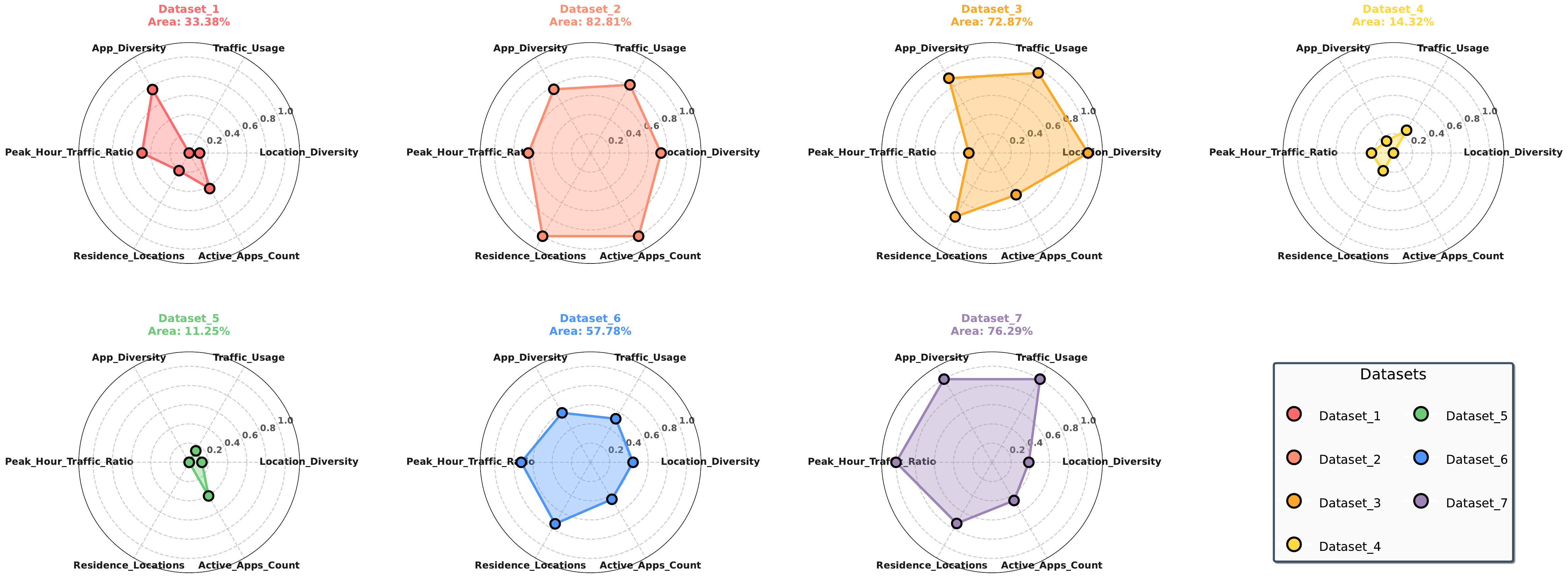}
\caption{The figure displays multi-dimensional user behavior characteristic comparisons 
across 7 datasets. Each 
subplot employs a polar coordinate system with six axes representing the six indicators: 
Location Diversity, Traffic Usage, App Diversity, Peak Hour Traffic Ratio, Residence 
Locations, and Active app Count. The colored area size indicates the overall activity 
level of user behavior. The legend in the lower right corner uses different colors to 
distinguish each dataset.}
\label{fig_dataset}
\end{figure*}

We evaluate U-MASK on seven real-world datasets for multivariate prediction of user location, app usage, and traffic volume~\cite{yu2018smartphone}. The characteristics of distribution shifts across these datasets are illustrated in Fig.~\ref{fig_dataset}.

\subsubsection{Multivariate prediction of user app usage, location, and network traffic}

For the baseline comparisons, we extend each deep learning model with conditional channels that incorporate user profiles derived from U-SCOPE profiles, which enables a controlled evaluation of how user semantics affect multivariate spatio-temporal prediction.
The baselines include: (i) Statistical forecasting methods, such as Historical Average (\textbf{HA}) and \textbf{ARIMA}~\cite{Xu2023MachineLF}; (ii) NLP-based models, with \textbf{Tempo}~\cite{Cao2023TEMPOPG} representing time-series dynamics through natural-language descriptions that are then used for prediction, and (iii) Spatio-temporal forecasting methods, including \textbf{CSDI}~\cite{Tashiro2021CSDICS} and \textbf{PatchTST}~\cite{Nie2022ATS}, which capture mobile traffic dynamics through autoregressive decomposition and spatio-temporal convolutions.


All methods are evaluated on multivariate forecasting tasks that jointly predict user location, app usage, and traffic volume, reflecting the coupled nature of mobile user behavior across space, time, and modality.
Due to the high variability of user mobility and app activity, we consider both short-term and long-term prediction settings to evaluate model performance under different temporal horizons.
We evaluate performance with root mean square error (RMSE) and mean absolute error (MAE).

\begin{table*}[t]
\centering
\caption{Short-term Traffic Prediction Performance Comparison. \textbf{\textcolor{posdelta}{Green}} values in $\Delta$ indicate performance improvement.}
\label{tab1}
\begingroup
\setlength{\tabcolsep}{2.2pt}
\renewcommand{\arraystretch}{0.851}
\scriptsize
\resizebox{0.95\linewidth}{!}{%
\begin{tabular}{l l c c c c c c c c c c c c c c}
\toprule
\rowcolor{headergray}
\textbf{Model} & \textbf{Setting}
& \multicolumn{2}{c}{\textbf{Dataset 1}}
& \multicolumn{2}{c}{\textbf{Dataset 2}}
& \multicolumn{2}{c}{\textbf{Dataset 3}}
& \multicolumn{2}{c}{\textbf{Dataset 4}}
& \multicolumn{2}{c}{\textbf{Dataset 5}}
& \multicolumn{2}{c}{\textbf{Dataset 6}}
& \multicolumn{2}{c}{\textbf{Dataset 7}} \\
\cmidrule(lr){3-4}\cmidrule(lr){5-6}\cmidrule(lr){7-8}\cmidrule(lr){9-10}\cmidrule(lr){11-12}\cmidrule(lr){13-14}\cmidrule(lr){15-16}
\rowcolor{headergray}
& & RMSE & MAE & RMSE & MAE & RMSE & MAE & RMSE & MAE & RMSE & MAE & RMSE & MAE & RMSE & MAE \\
\midrule

\textit{HA}    & w/o cond. & 0.3152 & 0.1034 & 1.0786 & 0.2027 & 0.1507 & 0.0513 & 0.1898 & 0.0591 & 0.1465 & 0.0483 & 0.7089 & 0.1170 & 0.1532 & 0.0454 \\
\textit{ARIMA} & w/o cond. & 0.1372 & 0.0560 & 0.5963 & 0.1202 & 0.1284 & 0.0464 & 0.1411 & 0.0465 & 0.1228 & 0.0408 & 0.1201 & 0.0452 & 0.1728 & 0.0565 \\
\midrule

\multirow{3}{*}{\textit{Tempo}}
& w/o cond. & 0.0174 & 0.0030 & 0.0138 & 0.0021 & 0.0178 & 0.0034 & 0.0282 & 0.0053 & 0.0187 & 0.0035 & 0.0308 & 0.0052 & 0.0172 & 0.0035 \\
& w/ cond.  & 0.0158 & 0.0022 & 0.0136 & 0.0023 & 0.0160 & 0.0021 & 0.0269 & 0.0043 & 0.0174 & 0.0033 & 0.0296 & 0.0040 & 0.0128 & 0.0028 \\
& $\Delta$  & \textcolor{posdelta}{+9.2\%}  & \textcolor{posdelta}{+26.7\%}
            & \textcolor{posdelta}{+1.4\%}  & \textcolor{negdelta}{-9.5\%}
            & \textcolor{posdelta}{+10.1\%} & \textcolor{posdelta}{+38.2\%}
            & \textcolor{posdelta}{+4.6\%}  & \textcolor{posdelta}{+18.9\%}
            & \textcolor{posdelta}{+7.0\%}  & \textcolor{posdelta}{+5.7\%}
            & \textcolor{posdelta}{+3.9\%}  & \textcolor{posdelta}{+23.1\%}
            & \textcolor{posdelta}{+25.6\%} & \textcolor{posdelta}{+20.0\%} \\
\midrule

\multirow{3}{*}{\textit{CSDI}}
& w/o cond. & 0.0593 & 0.0136 & 0.0675 & 0.0187 & 0.0549 & 0.0161 & 0.0597 & 0.0154 & 0.1367 & 0.0339 & 0.0561 & 0.0116 & 0.1130 & 0.0359 \\
& w/ cond.  & 0.0171 & 0.0030 & 0.0115 & 0.0040 & 0.0119 & 0.0028 & 0.0107 & 0.0028 & 0.0146 & 0.0068 & 0.0123 & 0.0032 & 0.0953 & 0.0235 \\
& $\Delta$  & \textcolor{posdelta}{+71.2\%} & \textcolor{posdelta}{+77.9\%}
            & \textcolor{posdelta}{+83.0\%} & \textcolor{posdelta}{+78.6\%}
            & \textcolor{posdelta}{+78.3\%} & \textcolor{posdelta}{+82.6\%}
            & \textcolor{posdelta}{+82.1\%} & \textcolor{posdelta}{+81.8\%}
            & \textcolor{posdelta}{+89.3\%} & \textcolor{posdelta}{+79.9\%}
            & \textcolor{posdelta}{+78.1\%} & \textcolor{posdelta}{+72.4\%}
            & \textcolor{posdelta}{+15.7\%} & \textcolor{posdelta}{+34.5\%} \\
\midrule

\multirow{3}{*}{\textit{DiT}}
& w/o cond. & 0.0550 & 0.0131 & 0.0580 & 0.0163 & 0.0752 & 0.0193 & 0.1282 & 0.0299 & 0.0800 & 0.0194 & 0.1546 & 0.0442 & 0.0793 & 0.0178 \\
& w/ cond.  & 0.0484 & 0.0118 & 0.0408 & 0.0100 & 0.0683 & 0.0188 & 0.0336 & 0.0089 & 0.0560 & 0.0133 & 0.0389 & 0.0127 & 0.0421 & 0.0114 \\
& $\Delta$  & \textcolor{posdelta}{+12.0\%} & \textcolor{posdelta}{+9.9\%}
            & \textcolor{posdelta}{+29.7\%} & \textcolor{posdelta}{+38.7\%}
            & \textcolor{posdelta}{+9.2\%}  & \textcolor{posdelta}{+2.6\%}
            & \textcolor{posdelta}{+73.8\%} & \textcolor{posdelta}{+70.2\%}
            & \textcolor{posdelta}{+30.0\%} & \textcolor{posdelta}{+31.4\%}
            & \textcolor{posdelta}{+74.8\%} & \textcolor{posdelta}{+71.3\%}
            & \textcolor{posdelta}{+46.9\%} & \textcolor{posdelta}{+36.0\%} \\
\midrule

\multirow{3}{*}{\textit{PatchTST}}
& w/o cond. & 0.0172 & 0.0028 & 0.0084 & 0.0025 & 0.0142 & 0.0033 & 0.0268 & 0.0053 & 0.0165 & 0.0036 & 0.0186 & 0.0047 & 0.0146 & 0.0032 \\
& w/ cond.  & 0.0176 & 0.0022 & 0.0087 & 0.0020 & 0.0135 & 0.0024 & 0.0263 & 0.0047 & 0.0160 & 0.0031 & 0.0171 & 0.0039 & 0.0142 & 0.0024 \\
& $\Delta$  & \textcolor{negdelta}{-2.3\%}  & \textcolor{posdelta}{+21.4\%}
            & \textcolor{negdelta}{-3.6\%}  & \textcolor{posdelta}{+20.0\%}
            & \textcolor{posdelta}{+4.9\%}  & \textcolor{posdelta}{+27.3\%}
            & \textcolor{posdelta}{+1.9\%}  & \textcolor{posdelta}{+11.3\%}
            & \textcolor{posdelta}{+3.0\%}  & \textcolor{posdelta}{+13.9\%}
            & \textcolor{posdelta}{+8.1\%}  & \textcolor{posdelta}{+17.0\%}
            & \textcolor{posdelta}{+2.7\%}  & \textcolor{posdelta}{+25.0\%} \\
\midrule

\rowcolor{oursbg}
\textbf{\textit{Ours}} & \textbf{U-MASK}
& \textbf{0.0042} & \textbf{0.0006}
& \textbf{0.0047} & \textbf{0.0028}
& \textbf{0.0023} & \textbf{0.0010}
& \textbf{0.0024} & \textbf{0.0005}
& \textbf{0.0034} & \textbf{0.0011}
& \textbf{0.0029} & \textbf{0.0005}
& \textbf{0.0040} & \textbf{0.0019} \\
\bottomrule
\end{tabular}%
}
\endgroup
\end{table*}

\begin{table*}[htbp]
\centering
\caption{Short-term App Prediction Performance. \textbf{\textcolor{posdelta}{Green}} values in $\Delta$ indicate performance improvement.}
\label{tab2}

\begingroup
\setlength{\tabcolsep}{2.2pt}
\renewcommand{\arraystretch}{0.851}
\scriptsize

\resizebox{0.95\linewidth}{!}{%
\begin{tabular}{l l c c c c c c c c c c c c c c}
\toprule
\rowcolor{headergray}
\textbf{Model} & \textbf{Setting}
& \multicolumn{2}{c}{\textbf{Dataset 1}}
& \multicolumn{2}{c}{\textbf{Dataset 2}}
& \multicolumn{2}{c}{\textbf{Dataset 3}}
& \multicolumn{2}{c}{\textbf{Dataset 4}}
& \multicolumn{2}{c}{\textbf{Dataset 5}}
& \multicolumn{2}{c}{\textbf{Dataset 6}}
& \multicolumn{2}{c}{\textbf{Dataset 7}} \\
\cmidrule(lr){3-4}\cmidrule(lr){5-6}\cmidrule(lr){7-8}\cmidrule(lr){9-10}\cmidrule(lr){11-12}\cmidrule(lr){13-14}\cmidrule(lr){15-16}
\rowcolor{headergray}
& & RMSE & MAE & RMSE & MAE & RMSE & MAE & RMSE & MAE & RMSE & MAE & RMSE & MAE & RMSE & MAE \\
\midrule

\textit{HA}    & w/o cond. & 0.2245 & 0.0644 & 0.1139 & 0.0496 & 0.2080 & 0.1197 & 0.0787 & 0.0343 & 0.2069 & 0.0478 & 0.1137 & 0.0441 & 0.1298 & 0.0493 \\
\textit{ARIMA} & w/o cond. & 0.1443 & 0.0555 & 0.0855 & 0.0394 & 0.0705 & 0.0283 & 0.0607 & 0.0259 & 0.2296 & 0.0479 & 0.0870 & 0.0356 & 0.1105 & 0.0440 \\
\midrule

\multirow{3}{*}{\textit{Tempo}}
& w/o cond. & 0.1136 & 0.0422 & 0.1110 & 0.0495 & 0.0959 & 0.0400 & 0.0908 & 0.0352 & 0.0834 & 0.0280 & 0.0887 & 0.0394 & 0.1047 & 0.0413 \\
& w/ cond.  & 0.1093 & 0.0371 & 0.1035 & 0.0404 & 0.0914 & 0.0351 & 0.0847 & 0.0300 & 0.0831 & 0.0259 & 0.0811 & 0.0311 & 0.1027 & 0.0368 \\
& $\Delta$  & \textcolor{posdelta}{+3.8\%}  & \textcolor{posdelta}{+12.1\%}
            & \textcolor{posdelta}{+6.8\%}  & \textcolor{posdelta}{+18.4\%}
            & \textcolor{posdelta}{+4.7\%}  & \textcolor{posdelta}{+12.3\%}
            & \textcolor{posdelta}{+6.7\%}  & \textcolor{posdelta}{+14.8\%}
            & \textcolor{posdelta}{+0.4\%}  & \textcolor{posdelta}{+7.5\%}
            & \textcolor{posdelta}{+8.6\%}  & \textcolor{posdelta}{+21.1\%}
            & \textcolor{posdelta}{+1.9\%}  & \textcolor{posdelta}{+10.9\%} \\
\midrule

\multirow{3}{*}{\textit{CSDI}}
& w/o cond. & 0.1182 & 0.0306 & 0.1070 & 0.0343 & 0.0853 & 0.0248 & 0.0662 & 0.0181 & 0.1367 & 0.0339 & 0.1002 & 0.0243 & 0.1130 & 0.0359 \\
& w/ cond.  & 0.1126 & 0.0276 & 0.0904 & 0.0232 & 0.0755 & 0.0171 & 0.0534 & 0.0114 & 0.0840 & 0.0196 & 0.0819 & 0.0182 & 0.0953 & 0.0235 \\
& $\Delta$  & \textcolor{posdelta}{+4.7\%}  & \textcolor{posdelta}{+9.8\%}
            & \textcolor{posdelta}{+15.5\%} & \textcolor{posdelta}{+32.4\%}
            & \textcolor{posdelta}{+11.5\%} & \textcolor{posdelta}{+31.0\%}
            & \textcolor{posdelta}{+19.3\%} & \textcolor{posdelta}{+37.0\%}
            & \textcolor{posdelta}{+38.6\%} & \textcolor{posdelta}{+42.2\%}
            & \textcolor{posdelta}{+18.3\%} & \textcolor{posdelta}{+25.1\%}
            & \textcolor{posdelta}{+15.7\%} & \textcolor{posdelta}{+34.5\%} \\
\midrule

\multirow{3}{*}{\textit{DiT}}
& w/o cond. & 0.0380 & 0.0109 & 0.0311 & 0.0121 & 0.0387 & 0.0132 & 0.0387 & 0.0117 & 0.0289 & 0.0083 & 0.0714 & 0.0205 & 0.0284 & 0.0090 \\
& w/ cond.  & 0.0317 & 0.0101 & 0.0202 & 0.0080 & 0.0297 & 0.0117 & 0.0204 & 0.0075 & 0.0364 & 0.0096 & 0.0295 & 0.0103 & 0.0212 & 0.0077 \\
& $\Delta$  & \textcolor{posdelta}{+16.6\%} & \textcolor{posdelta}{+7.3\%}
            & \textcolor{posdelta}{+35.0\%} & \textcolor{posdelta}{+33.9\%}
            & \textcolor{posdelta}{+23.3\%} & \textcolor{posdelta}{+11.4\%}
            & \textcolor{posdelta}{+47.3\%} & \textcolor{posdelta}{+35.9\%}
            & \textcolor{negdelta}{-26.0\%} & \textcolor{negdelta}{-15.7\%}
            & \textcolor{posdelta}{+58.7\%} & \textcolor{posdelta}{+49.8\%}
            & \textcolor{posdelta}{+25.4\%} & \textcolor{posdelta}{+14.4\%} \\
\midrule

\multirow{3}{*}{\textit{PatchTST}}
& w/o cond. & 0.0977 & 0.0401 & 0.0860 & 0.0416 & 0.0786 & 0.0348 & 0.0647 & 0.0284 & 0.0835 & 0.0293 & 0.0810 & 0.0335 & 0.0914 & 0.0413 \\
& w/ cond.  & 0.0889 & 0.0253 & 0.0726 & 0.0209 & 0.0675 & 0.0197 & 0.0527 & 0.0180 & 0.0788 & 0.0190 & 0.0730 & 0.0189 & 0.0799 & 0.0244 \\
& $\Delta$  & \textcolor{posdelta}{+9.0\%}  & \textcolor{posdelta}{+36.9\%}
            & \textcolor{posdelta}{+15.6\%} & \textcolor{posdelta}{+49.8\%}
            & \textcolor{posdelta}{+14.1\%} & \textcolor{posdelta}{+43.4\%}
            & \textcolor{posdelta}{+18.5\%} & \textcolor{posdelta}{+36.6\%}
            & \textcolor{posdelta}{+5.6\%}  & \textcolor{posdelta}{+35.2\%}
            & \textcolor{posdelta}{+9.9\%}  & \textcolor{posdelta}{+43.6\%}
            & \textcolor{posdelta}{+12.6\%} & \textcolor{posdelta}{+40.9\%} \\
\midrule

\rowcolor{oursbg}
\textbf{\textit{Ours}} & \textbf{U-MASK}
& \textbf{0.0129} & \textbf{0.0040}
& \textbf{0.0076} & \textbf{0.0028}
& \textbf{0.0098} & \textbf{0.0025}
& \textbf{0.0098} & \textbf{0.0031}
& \textbf{0.0106} & \textbf{0.0029}
& \textbf{0.0096} & \textbf{0.0028}
& \textbf{0.0113} & \textbf{0.0039} \\
\bottomrule
\end{tabular}%
}
\endgroup
\end{table*}

\begin{table*}[htbp]
\centering
\caption{Short-term Location Prediction Performance. \textbf{\textcolor{posdelta}{Green}} values in $\Delta$ indicate performance improvement.}
\label{tab3}

\begingroup
\setlength{\tabcolsep}{2.2pt}
\renewcommand{\arraystretch}{0.851}
\scriptsize

\resizebox{0.95\linewidth}{!}{%
\begin{tabular}{l l c c c c c c c c c c c c c c}
\toprule
\rowcolor{headergray}
\textbf{Model} & \textbf{Setting}
& \multicolumn{2}{c}{\textbf{Dataset 1}}
& \multicolumn{2}{c}{\textbf{Dataset 2}}
& \multicolumn{2}{c}{\textbf{Dataset 3}}
& \multicolumn{2}{c}{\textbf{Dataset 4}}
& \multicolumn{2}{c}{\textbf{Dataset 5}}
& \multicolumn{2}{c}{\textbf{Dataset 6}}
& \multicolumn{2}{c}{\textbf{Dataset 7}} \\
\cmidrule(lr){3-4}\cmidrule(lr){5-6}\cmidrule(lr){7-8}\cmidrule(lr){9-10}\cmidrule(lr){11-12}\cmidrule(lr){13-14}\cmidrule(lr){15-16}
\rowcolor{headergray}
& & RMSE & MAE & RMSE & MAE & RMSE & MAE & RMSE & MAE & RMSE & MAE & RMSE & MAE & RMSE & MAE \\
\midrule

\textit{HA}    & w/o cond. & 0.2369 & 0.1139 & 0.2080 & 0.1197 & 0.1381 & 0.0808 & 0.1538 & 0.0732 & 0.3783 & 0.1624 & 0.2663 & 0.1230 & 0.2186 & 0.1238 \\
\textit{ARIMA} & w/o cond. & 0.1660 & 0.0955 & 0.1897 & 0.1102 & 0.1321 & 0.0801 & 0.1259 & 0.0664 & 0.2389 & 0.1344 & 0.2066 & 0.0915 & 0.1998 & 0.1209 \\
\midrule

\multirow{3}{*}{\textit{Tempo}}
& w/o cond. & 0.1726 & 0.0842 & 0.2164 & 0.1202 & 0.1389 & 0.0817 & 0.1567 & 0.0865 & 0.2348 & 0.1287 & 0.1921 & 0.0948 & 0.1814 & 0.1065 \\
& w/ cond.  & 0.1672 & 0.0777 & 0.2014 & 0.1043 & 0.1319 & 0.0766 & 0.1462 & 0.0755 & 0.2334 & 0.1198 & 0.1796 & 0.0840 & 0.1792 & 0.1021 \\
& $\Delta$  & \textcolor{posdelta}{+3.1\%}  & \textcolor{posdelta}{+7.6\%}
            & \textcolor{posdelta}{+6.9\%}  & \textcolor{posdelta}{+13.2\%}
            & \textcolor{posdelta}{+5.3\%}  & \textcolor{posdelta}{+6.3\%}
            & \textcolor{posdelta}{+6.7\%}  & \textcolor{posdelta}{+12.7\%}
            & \textcolor{posdelta}{+0.2\%}  & \textcolor{posdelta}{+7.0\%}
            & \textcolor{posdelta}{+7.3\%}  & \textcolor{posdelta}{+12.7\%}
            & \textcolor{posdelta}{+1.7\%}  & \textcolor{posdelta}{+4.3\%} \\
\midrule

\multirow{3}{*}{\textit{CSDI}}
& w/o cond. & 0.1772 & 0.0715 & 0.2031 & 0.0766 & 0.1301 & 0.0509 & 0.1362 & 0.0449 & 0.2388 & 0.0887 & 0.1816 & 0.0618 & 0.1770 & 0.0693 \\
& w/ cond.  & 0.1591 & 0.0548 & 0.1595 & 0.0540 & 0.1211 & 0.0446 & 0.1189 & 0.0356 & 0.2110 & 0.0657 & 0.1706 & 0.0500 & 0.1619 & 0.0601 \\
& $\Delta$  & \textcolor{posdelta}{+10.2\%} & \textcolor{posdelta}{+23.4\%}
            & \textcolor{posdelta}{+21.5\%} & \textcolor{posdelta}{+29.5\%}
            & \textcolor{posdelta}{+6.9\%}  & \textcolor{posdelta}{+12.4\%}
            & \textcolor{posdelta}{+12.7\%} & \textcolor{posdelta}{+20.7\%}
            & \textcolor{posdelta}{+11.6\%} & \textcolor{posdelta}{+25.9\%}
            & \textcolor{posdelta}{+6.1\%}  & \textcolor{posdelta}{+19.1\%}
            & \textcolor{posdelta}{+8.5\%}  & \textcolor{posdelta}{+13.3\%} \\
\midrule

\multirow{3}{*}{\textit{DiT}}
& w/o cond. & 0.0398 & 0.0112 & 0.0800 & 0.0203 & 0.0545 & 0.0146 & 0.0904 & 0.0221 & 0.0624 & 0.0147 & 0.1081 & 0.0335 & 0.0612 & 0.0159 \\
& w/ cond.  & 0.0380 & 0.0104 & 0.0290 & 0.0069 & 0.0393 & 0.0118 & 0.0348 & 0.0082 & 0.0449 & 0.0109 & 0.0417 & 0.0115 & 0.0388 & 0.0102 \\
& $\Delta$  & \textcolor{posdelta}{+4.5\%}  & \textcolor{posdelta}{+7.1\%}
            & \textcolor{posdelta}{+63.8\%} & \textcolor{posdelta}{+66.0\%}
            & \textcolor{posdelta}{+27.9\%} & \textcolor{posdelta}{+19.2\%}
            & \textcolor{posdelta}{+61.5\%} & \textcolor{posdelta}{+62.9\%}
            & \textcolor{posdelta}{+28.0\%} & \textcolor{posdelta}{+25.9\%}
            & \textcolor{posdelta}{+61.4\%} & \textcolor{posdelta}{+65.7\%}
            & \textcolor{posdelta}{+36.6\%} & \textcolor{posdelta}{+35.8\%} \\
\midrule

\multirow{3}{*}{\textit{PatchTST}}
& w/o cond. & 0.1553 & 0.0895 & 0.1919 & 0.1152 & 0.1243 & 0.0774 & 0.1263 & 0.0724 & 0.2110 & 0.1248 & 0.1816 & 0.0969 & 0.1776 & 0.1163 \\
& w/ cond.  & 0.1411 & 0.0614 & 0.1568 & 0.0699 & 0.1089 & 0.0494 & 0.1049 & 0.0490 & 0.1846 & 0.0894 & 0.1573 & 0.0594 & 0.1496 & 0.0727 \\
& $\Delta$  & \textcolor{posdelta}{+9.1\%}  & \textcolor{posdelta}{+31.4\%}
            & \textcolor{posdelta}{+18.3\%} & \textcolor{posdelta}{+39.3\%}
            & \textcolor{posdelta}{+12.4\%} & \textcolor{posdelta}{+36.2\%}
            & \textcolor{posdelta}{+16.9\%} & \textcolor{posdelta}{+32.3\%}
            & \textcolor{posdelta}{+12.5\%} & \textcolor{posdelta}{+28.4\%}
            & \textcolor{posdelta}{+13.4\%} & \textcolor{posdelta}{+38.7\%}
            & \textcolor{posdelta}{+15.8\%} & \textcolor{posdelta}{+37.5\%} \\
\midrule

\rowcolor{oursbg}
\textbf{\textit{Ours}} & \textbf{U-MASK}
& \textbf{0.0145} & \textbf{0.0041}
& \textbf{0.0224} & \textbf{0.0070}
& \textbf{0.0097} & \textbf{0.0028}
& \textbf{0.0090} & \textbf{0.0028}
& \textbf{0.0105} & \textbf{0.0031}
& \textbf{0.0098} & \textbf{0.0030}
& \textbf{0.0091} & \textbf{0.0032} \\
\bottomrule
\end{tabular}%
}
\endgroup
\end{table*}

\begin{table*}[htbp]
\centering
\caption{Long-term Traffic Prediction Performance. \textbf{\textcolor{posdelta}{Green}} values in $\Delta$ indicate performance improvement.}
\label{tab4}

\begingroup
\setlength{\tabcolsep}{2.2pt}
\renewcommand{\arraystretch}{0.851}
\scriptsize

\resizebox{0.95\linewidth}{!}{%
\begin{tabular}{l l c c c c c c c c c c c c c c}
\toprule
\rowcolor{headergray}
\textbf{Model} & \textbf{Setting}
& \multicolumn{2}{c}{\textbf{Dataset 1}}
& \multicolumn{2}{c}{\textbf{Dataset 2}}
& \multicolumn{2}{c}{\textbf{Dataset 3}}
& \multicolumn{2}{c}{\textbf{Dataset 4}}
& \multicolumn{2}{c}{\textbf{Dataset 5}}
& \multicolumn{2}{c}{\textbf{Dataset 6}}
& \multicolumn{2}{c}{\textbf{Dataset 7}} \\
\cmidrule(lr){3-4}\cmidrule(lr){5-6}\cmidrule(lr){7-8}\cmidrule(lr){9-10}\cmidrule(lr){11-12}\cmidrule(lr){13-14}\cmidrule(lr){15-16}
\rowcolor{headergray}
& & RMSE & MAE & RMSE & MAE & RMSE & MAE & RMSE & MAE & RMSE & MAE & RMSE & MAE & RMSE & MAE \\
\midrule

\textit{HA}    & w/o cond. & 0.7648 & 0.3571 & 1.0825 & 0.6119 & 0.0174 & 0.0030 & 0.0307 & 0.0067 & 0.0645 & 0.0126 & 0.0352 & 0.0075 & 0.0237 & 0.0062 \\
\textit{ARIMA} & w/o cond. & 0.3927 & 0.1764 & 0.5578 & 0.3161 & 1.0839 & 0.5297 & 1.1439 & 0.4748 & 0.9236 & 0.4844 & 0.8279 & 0.3878 & 1.1880 & 0.4697 \\
\midrule

\multirow{3}{*}{\textit{Tempo}}
& w/o cond. & 0.0181 & 0.0032 & 0.0076 & 0.0027 & 0.0124 & 0.0020 & 0.0246 & 0.0041 & 0.0151 & 0.0034 & 0.0125 & 0.0037 & 0.0089 & 0.0028 \\
& w/ cond.  & 0.0176 & 0.0029 & 0.0061 & 0.0021 & 0.0120 & 0.0022 & 0.0241 & 0.0032 & 0.0145 & 0.0025 & 0.0105 & 0.0025 & 0.0049 & 0.0016 \\
& $\Delta$  & \textcolor{posdelta}{+2.8\%}  & \textcolor{posdelta}{+9.4\%}
            & \textcolor{posdelta}{+19.7\%} & \textcolor{posdelta}{+22.2\%}
            & \textcolor{posdelta}{+3.2\%}  & \textcolor{negdelta}{-10.0\%}
            & \textcolor{posdelta}{+2.0\%}  & \textcolor{posdelta}{+22.0\%}
            & \textcolor{posdelta}{+4.0\%}  & \textcolor{posdelta}{+26.5\%}
            & \textcolor{posdelta}{+16.0\%} & \textcolor{posdelta}{+32.4\%}
            & \textcolor{posdelta}{+44.9\%} & \textcolor{posdelta}{+42.9\%} \\
\midrule

\multirow{3}{*}{\textit{CSDI}}
& w/o cond. & 0.0779 & 0.0237 & 0.0788 & 0.0195 & 0.0752 & 0.0257 & 0.0685 & 0.0172 & 0.1306 & 0.0305 & 0.0830 & 0.0249 & 0.0714 & 0.0171 \\
& w/ cond.  & 0.0221 & 0.0094 & 0.0167 & 0.0042 & 0.0151 & 0.0042 & 0.0252 & 0.0135 & 0.0150 & 0.0037 & 0.0225 & 0.0053 & 0.0124 & 0.0035 \\
& $\Delta$  & \textcolor{posdelta}{+71.6\%} & \textcolor{posdelta}{+60.3\%}
            & \textcolor{posdelta}{+78.8\%} & \textcolor{posdelta}{+78.5\%}
            & \textcolor{posdelta}{+79.9\%} & \textcolor{posdelta}{+83.7\%}
            & \textcolor{posdelta}{+63.2\%} & \textcolor{posdelta}{+21.5\%}
            & \textcolor{posdelta}{+88.5\%} & \textcolor{posdelta}{+87.9\%}
            & \textcolor{posdelta}{+72.9\%} & \textcolor{posdelta}{+78.7\%}
            & \textcolor{posdelta}{+82.6\%} & \textcolor{posdelta}{+79.5\%} \\
\midrule

\multirow{3}{*}{\textit{DiT}}
& w/o cond. & 0.2299 & 0.0574 & 0.0988 & 0.0226 & 0.1767 & 0.0438 & 0.2478 & 0.0682 & 0.2286 & 0.0541 & 0.1887 & 0.0433 & 0.1912 & 0.0505 \\
& w/ cond.  & 0.0552 & 0.0152 & 0.0357 & 0.0113 & 0.0431 & 0.0130 & 0.0723 & 0.0169 & 0.0744 & 0.0198 & 0.0442 & 0.0112 & 0.0418 & 0.0125 \\
& $\Delta$  & \textcolor{posdelta}{+76.0\%} & \textcolor{posdelta}{+73.5\%}
            & \textcolor{posdelta}{+63.9\%} & \textcolor{posdelta}{+50.0\%}
            & \textcolor{posdelta}{+75.6\%} & \textcolor{posdelta}{+70.3\%}
            & \textcolor{posdelta}{+70.8\%} & \textcolor{posdelta}{+75.2\%}
            & \textcolor{posdelta}{+67.5\%} & \textcolor{posdelta}{+63.4\%}
            & \textcolor{posdelta}{+76.6\%} & \textcolor{posdelta}{+74.1\%}
            & \textcolor{posdelta}{+78.1\%} & \textcolor{posdelta}{+75.2\%} \\
\midrule

\multirow{3}{*}{\textit{PatchTST}}
& w/o cond. & 0.0184 & 0.0031 & 0.0144 & 0.0029 & 0.0184 & 0.0036 & 0.0284 & 0.0053 & 0.0213 & 0.0040 & 0.0309 & 0.0053 & 0.0320 & 0.0056 \\
& w/ cond.  & 0.0184 & 0.0023 & 0.0138 & 0.0019 & 0.0172 & 0.0024 & 0.0275 & 0.0042 & 0.0225 & 0.0038 & 0.0311 & 0.0045 & 0.0304 & 0.0046 \\
& $\Delta$  & \textcolor{posdelta}{+0.0\%}  & \textcolor{posdelta}{+25.8\%}
            & \textcolor{posdelta}{+4.2\%}  & \textcolor{posdelta}{+34.5\%}
            & \textcolor{posdelta}{+6.5\%}  & \textcolor{posdelta}{+33.3\%}
            & \textcolor{posdelta}{+3.2\%}  & \textcolor{posdelta}{+20.8\%}
            & \textcolor{negdelta}{-5.6\%}  & \textcolor{posdelta}{+5.0\%}
            & \textcolor{negdelta}{-0.6\%}  & \textcolor{posdelta}{+15.1\%}
            & \textcolor{posdelta}{+5.0\%}  & \textcolor{posdelta}{+17.9\%} \\
\midrule

\rowcolor{oursbg}
\textbf{\textit{Ours}} & \textbf{U-MASK}
& \textbf{0.0054} & \textbf{0.0010}
& \textbf{0.0070} & \textbf{0.0052}
& \textbf{0.0043} & \textbf{0.0008}
& \textbf{0.0047} & \textbf{0.0021}
& \textbf{0.0048} & \textbf{0.0013}
& \textbf{0.0056} & \textbf{0.0012}
& \textbf{0.0054} & \textbf{0.0016} \\
\bottomrule
\end{tabular}%
}
\endgroup
\end{table*}

\begin{table*}[htbp]
\centering
\caption{Long-term App Prediction Performance. \textbf{\textcolor{posdelta}{Green}} values in $\Delta$ indicate performance improvement.}
\label{tab5}

\begingroup
\setlength{\tabcolsep}{2.2pt}
\renewcommand{\arraystretch}{0.851}
\scriptsize

\resizebox{0.95\linewidth}{!}{%
\begin{tabular}{l l c c c c c c c c c c c c c c}
\toprule
\rowcolor{headergray}
\textbf{Model} & \textbf{Setting}
& \multicolumn{2}{c}{\textbf{Dataset 1}}
& \multicolumn{2}{c}{\textbf{Dataset 2}}
& \multicolumn{2}{c}{\textbf{Dataset 3}}
& \multicolumn{2}{c}{\textbf{Dataset 4}}
& \multicolumn{2}{c}{\textbf{Dataset 5}}
& \multicolumn{2}{c}{\textbf{Dataset 6}}
& \multicolumn{2}{c}{\textbf{Dataset 7}} \\
\cmidrule(lr){3-4}\cmidrule(lr){5-6}\cmidrule(lr){7-8}\cmidrule(lr){9-10}\cmidrule(lr){11-12}\cmidrule(lr){13-14}\cmidrule(lr){15-16}
\rowcolor{headergray}
& & RMSE & MAE & RMSE & MAE & RMSE & MAE & RMSE & MAE & RMSE & MAE & RMSE & MAE & RMSE & MAE \\
\midrule

\textit{HA}    & w/o cond. & 0.3898 & 0.1300 & 0.3421 & 0.1537 & 0.5351 & 0.1476 & 0.2688 & 0.1076 & 0.2772 & 0.1139 & 1.0564 & 0.1991 & 0.3154 & 0.1336 \\
\textit{ARIMA} & w/o cond. & 0.0156 & 0.0015 & 0.1381 & 0.0602 & 0.1022 & 0.0424 & 0.1042 & 0.0350 & 0.1149 & 0.0400 & 0.0963 & 0.0362 & 0.1561 & 0.0562 \\
\midrule

\multirow{3}{*}{\textit{Tempo}}
& w/o cond. & 0.0961 & 0.0390 & 0.0783 & 0.0307 & 0.0690 & 0.0236 & 0.0533 & 0.0189 & 0.0820 & 0.0269 & 0.0767 & 0.0280 & 0.0856 & 0.0341 \\
& w/ cond.  & 0.0960 & 0.0353 & 0.0737 & 0.0247 & 0.0673 & 0.0200 & 0.0508 & 0.0151 & 0.0785 & 0.0195 & 0.0743 & 0.0236 & 0.0750 & 0.0256 \\
& $\Delta$  & \textcolor{posdelta}{+0.1\%}  & \textcolor{posdelta}{+9.5\%}
            & \textcolor{posdelta}{+5.9\%}  & \textcolor{posdelta}{+19.5\%}
            & \textcolor{posdelta}{+2.5\%}  & \textcolor{posdelta}{+15.3\%}
            & \textcolor{posdelta}{+4.7\%}  & \textcolor{posdelta}{+20.1\%}
            & \textcolor{posdelta}{+4.3\%}  & \textcolor{posdelta}{+27.5\%}
            & \textcolor{posdelta}{+3.1\%}  & \textcolor{posdelta}{+15.7\%}
            & \textcolor{posdelta}{+12.4\%} & \textcolor{posdelta}{+24.9\%} \\
\midrule

\multirow{3}{*}{\textit{CSDI}}
& w/o cond. & 0.1280 & 0.0397 & 0.1271 & 0.0399 & 0.1121 & 0.0417 & 0.1072 & 0.0330 & 0.1519 & 0.0428 & 0.1171 & 0.0401 & 0.1302 & 0.0421 \\
& w/ cond.  & 0.1225 & 0.0387 & 0.1159 & 0.0371 & 0.2569 & 0.1203 & 0.0929 & 0.0334 & 0.0965 & 0.0243 & 0.0913 & 0.0273 & 0.1190 & 0.0351 \\
& $\Delta$  & \textcolor{posdelta}{+4.3\%}  & \textcolor{posdelta}{+2.5\%}
            & \textcolor{posdelta}{+8.8\%}  & \textcolor{posdelta}{+7.0\%}
            & \textcolor{negdelta}{-129.2\%} & \textcolor{negdelta}{-188.5\%}
            & \textcolor{posdelta}{+13.3\%} & \textcolor{negdelta}{-1.2\%}
            & \textcolor{posdelta}{+36.5\%} & \textcolor{posdelta}{+43.2\%}
            & \textcolor{posdelta}{+22.0\%} & \textcolor{posdelta}{+31.9\%}
            & \textcolor{posdelta}{+8.6\%}  & \textcolor{posdelta}{+16.6\%} \\
\midrule

\multirow{3}{*}{\textit{DiT}}
& w/o cond. & 0.0779 & 0.0209 & 0.0492 & 0.0144 & 0.0678 & 0.0209 & 0.0810 & 0.0228 & 0.0733 & 0.0161 & 0.0661 & 0.0163 & 0.0790 & 0.0217 \\
& w/ cond.  & 0.0343 & 0.0116 & 0.0297 & 0.0107 & 0.0255 & 0.0096 & 0.0302 & 0.0100 & 0.0452 & 0.0151 & 0.0262 & 0.0094 & 0.0302 & 0.0109 \\
& $\Delta$  & \textcolor{posdelta}{+56.0\%} & \textcolor{posdelta}{+44.5\%}
            & \textcolor{posdelta}{+39.6\%} & \textcolor{posdelta}{+25.7\%}
            & \textcolor{posdelta}{+62.4\%} & \textcolor{posdelta}{+54.1\%}
            & \textcolor{posdelta}{+62.7\%} & \textcolor{posdelta}{+56.1\%}
            & \textcolor{posdelta}{+38.3\%} & \textcolor{posdelta}{+6.2\%}
            & \textcolor{posdelta}{+60.4\%} & \textcolor{posdelta}{+42.3\%}
            & \textcolor{posdelta}{+61.8\%} & \textcolor{posdelta}{+49.8\%} \\
\midrule

\multirow{3}{*}{\textit{PatchTST}}
& w/o cond. & 0.1141 & 0.0432 & 0.1129 & 0.0513 & 0.0997 & 0.0423 & 0.0904 & 0.0335 & 0.0867 & 0.0303 & 0.0900 & 0.0387 & 0.1111 & 0.0474 \\
& w/ cond.  & 0.1071 & 0.0357 & 0.1015 & 0.0358 & 0.0897 & 0.0307 & 0.0855 & 0.0274 & 0.0871 & 0.0298 & 0.0827 & 0.0313 & 0.1061 & 0.0409 \\
& $\Delta$  & \textcolor{posdelta}{+6.1\%}  & \textcolor{posdelta}{+17.4\%}
            & \textcolor{posdelta}{+10.1\%} & \textcolor{posdelta}{+30.2\%}
            & \textcolor{posdelta}{+10.0\%} & \textcolor{posdelta}{+27.4\%}
            & \textcolor{posdelta}{+5.4\%}  & \textcolor{posdelta}{+18.2\%}
            & \textcolor{negdelta}{-0.5\%}  & \textcolor{posdelta}{+1.7\%}
            & \textcolor{posdelta}{+8.1\%}  & \textcolor{posdelta}{+19.1\%}
            & \textcolor{posdelta}{+4.5\%}  & \textcolor{posdelta}{+13.7\%} \\
\midrule

\rowcolor{oursbg}
\textbf{\textit{Ours}} & \textbf{U-MASK}
& \textbf{0.0201} & \textbf{0.0073}
& \textbf{0.0138} & \textbf{0.0041}
& \textbf{0.0158} & \textbf{0.0052}
& \textbf{0.0166} & \textbf{0.0070}
& \textbf{0.0172} & \textbf{0.0061}
& \textbf{0.0172} & \textbf{0.0067}
& \textbf{0.0168} & \textbf{0.0066} \\
\bottomrule
\end{tabular}%
}
\endgroup
\end{table*}

\begin{table*}[t]
\centering
\caption{Long-term Location Prediction Performance. \textbf{\textcolor{posdelta}{Green}} values in $\Delta$ indicate performance improvement.}
\label{tab6}

\begingroup
\setlength{\tabcolsep}{2.2pt}
\renewcommand{\arraystretch}{0.851}
\scriptsize

\resizebox{0.95\linewidth}{!}{%
\begin{tabular}{l l c c c c c c c c c c c c c c}
\toprule
\rowcolor{headergray}
\textbf{Model} & \textbf{Setting}
& \multicolumn{2}{c}{\textbf{Dataset 1}}
& \multicolumn{2}{c}{\textbf{Dataset 2}}
& \multicolumn{2}{c}{\textbf{Dataset 3}}
& \multicolumn{2}{c}{\textbf{Dataset 4}}
& \multicolumn{2}{c}{\textbf{Dataset 5}}
& \multicolumn{2}{c}{\textbf{Dataset 6}}
& \multicolumn{2}{c}{\textbf{Dataset 7}} \\
\cmidrule(lr){3-4}\cmidrule(lr){5-6}\cmidrule(lr){7-8}\cmidrule(lr){9-10}\cmidrule(lr){11-12}\cmidrule(lr){13-14}\cmidrule(lr){15-16}
\rowcolor{headergray}
& & RMSE & MAE & RMSE & MAE & RMSE & MAE & RMSE & MAE & RMSE & MAE & RMSE & MAE & RMSE & MAE \\
\midrule

\textit{HA}    & w/o cond. & 0.6957 & 0.2552 & 0.9168 & 0.3659 & 0.4562 & 0.2318 & 0.4485 & 0.2071 & 0.6919 & 0.4077 & 0.5298 & 0.2388 & 1.3408 & 0.3968 \\
\textit{ARIMA} & w/o cond. & 0.1660 & 0.0955 & 0.1897 & 0.1102 & 0.1322 & 0.0801 & 0.1259 & 0.0664 & 0.2389 & 0.1344 & 0.2066 & 0.0915 & 0.1998 & 0.1209 \\
\midrule

\multirow{3}{*}{\textit{Tempo}}
& w/o cond. & 0.1526 & 0.0809 & 0.1557 & 0.0757 & 0.1060 & 0.0557 & 0.1073 & 0.0502 & 0.1881 & 0.0996 & 0.1572 & 0.0740 & 0.0248 & 0.0917 \\
& w/ cond.  & 0.1395 & 0.0694 & 0.1531 & 0.0732 & 0.1079 & 0.0539 & 0.1033 & 0.0466 & 0.1793 & 0.0858 & 0.1494 & 0.0665 & 0.1393 & 0.0770 \\
& $\Delta$  & \textcolor{posdelta}{+8.6\%}  & \textcolor{posdelta}{+14.2\%}
            & \textcolor{posdelta}{+1.7\%}  & \textcolor{posdelta}{+3.3\%}
            & \textcolor{negdelta}{-1.8\%}  & \textcolor{posdelta}{+3.2\%}
            & \textcolor{posdelta}{+3.7\%}  & \textcolor{posdelta}{+7.2\%}
            & \textcolor{posdelta}{+4.7\%}  & \textcolor{posdelta}{+13.9\%}
            & \textcolor{posdelta}{+5.0\%}  & \textcolor{posdelta}{+10.1\%}
            & \textcolor{negdelta}{-461.7\%} & \textcolor{posdelta}{+16.0\%} \\
\midrule

\multirow{3}{*}{\textit{CSDI}}
& w/o cond. & 0.1918 & 0.0877 & 0.2569 & 0.1203 & 0.1670 & 0.0860 & 0.1727 & 0.0769 & 0.3015 & 0.1331 & 0.2353 & 0.1039 & 0.2380 & 0.1212 \\
& w/ cond.  & 0.1794 & 0.0614 & 0.2184 & 0.0881 & 0.1514 & 0.0707 & 0.1543 & 0.0570 & 0.2548 & 0.0972 & 0.2003 & 0.0773 & 0.2034 & 0.0891 \\
& $\Delta$  & \textcolor{posdelta}{+6.5\%}  & \textcolor{posdelta}{+30.0\%}
            & \textcolor{posdelta}{+15.0\%} & \textcolor{posdelta}{+26.8\%}
            & \textcolor{posdelta}{+9.3\%}  & \textcolor{posdelta}{+17.8\%}
            & \textcolor{posdelta}{+10.7\%} & \textcolor{posdelta}{+25.9\%}
            & \textcolor{posdelta}{+15.5\%} & \textcolor{posdelta}{+27.0\%}
            & \textcolor{posdelta}{+14.9\%} & \textcolor{posdelta}{+25.6\%}
            & \textcolor{posdelta}{+14.5\%} & \textcolor{posdelta}{+26.5\%} \\
\midrule

\multirow{3}{*}{\textit{DiT}}
& w/o cond. & 0.1366 & 0.0362 & 0.1390 & 0.0293 & 0.1275 & 0.0369 & 0.1601 & 0.0443 & 0.1475 & 0.0350 & 0.1321 & 0.0313 & 0.1354 & 0.0395 \\
& w/ cond.  & 0.0406 & 0.0114 & 0.0420 & 0.0088 & 0.0345 & 0.0098 & 0.0446 & 0.0115 & 0.0568 & 0.0162 & 0.0363 & 0.0094 & 0.0342 & 0.0099 \\
& $\Delta$  & \textcolor{posdelta}{+70.3\%} & \textcolor{posdelta}{+68.5\%}
            & \textcolor{posdelta}{+69.8\%} & \textcolor{posdelta}{+70.0\%}
            & \textcolor{posdelta}{+72.9\%} & \textcolor{posdelta}{+73.4\%}
            & \textcolor{posdelta}{+72.1\%} & \textcolor{posdelta}{+74.0\%}
            & \textcolor{posdelta}{+61.5\%} & \textcolor{posdelta}{+53.7\%}
            & \textcolor{posdelta}{+72.5\%} & \textcolor{posdelta}{+70.0\%}
            & \textcolor{posdelta}{+74.7\%} & \textcolor{posdelta}{+74.9\%} \\
\midrule

\multirow{3}{*}{\textit{PatchTST}}
& w/o cond. & 0.1756 & 0.0897 & 0.2247 & 0.1294 & 0.1403 & 0.0866 & 0.1563 & 0.0857 & 0.2467 & 0.1411 & 0.1990 & 0.1023 & 0.1953 & 0.1220 \\
& w/ cond.  & 0.1678 & 0.0811 & 0.2017 & 0.1045 & 0.1302 & 0.0750 & 0.1451 & 0.0782 & 0.2359 & 0.1328 & 0.1819 & 0.0862 & 0.1830 & 0.1119 \\
& $\Delta$  & \textcolor{posdelta}{+4.4\%}  & \textcolor{posdelta}{+9.6\%}
            & \textcolor{posdelta}{+10.2\%} & \textcolor{posdelta}{+19.2\%}
            & \textcolor{posdelta}{+7.2\%}  & \textcolor{posdelta}{+13.4\%}
            & \textcolor{posdelta}{+7.2\%}  & \textcolor{posdelta}{+8.8\%}
            & \textcolor{posdelta}{+4.4\%}  & \textcolor{posdelta}{+5.9\%}
            & \textcolor{posdelta}{+8.6\%}  & \textcolor{posdelta}{+15.7\%}
            & \textcolor{posdelta}{+6.3\%}  & \textcolor{posdelta}{+8.3\%} \\
\midrule

\rowcolor{oursbg}
\textbf{\textit{Ours}} & \textbf{U-MASK}
& \textbf{0.0241} & \textbf{0.0114}
& \textbf{0.0390} & \textbf{0.0171}
& \textbf{0.0166} & \textbf{0.0069}
& \textbf{0.0161} & \textbf{0.0061}
& \textbf{0.0177} & \textbf{0.0070}
& \textbf{0.0219} & \textbf{0.0100}
& \textbf{0.0162} & \textbf{0.0077} \\
\bottomrule
\end{tabular}%
}
\endgroup
\end{table*}

Experiments are conducted on seven real-world mobile datasets~\cite{yu2018smartphone} for both short- and long-term multivariate prediction of user location, app usage, and traffic volume (Tables~\ref{tab1}–\ref{tab6}).
These seven datasets cover the period from April 20 to 26, 2016 and include records for 3000 users. 
Each user has 192 time steps at 15 minute intervals and a spatial resolution of 500m.
Across all datasets and prediction settings, our method consistently outperforms all baselines, including statistical models, NLP-based approaches, and advanced spatio-temporal architectures, without dataset-specific tuning.
The advantage arises from combining user-aware conditioning and task-adaptive feature selection. User profiles generated by U-SCOPE enable personalization to individual behavior, while the masking mechanism selectively exposes the most relevant temporal segments and spatial regions for each task. This approach accurately models intricate interdependencies between location, app, and traffic.
In contrast, statistical models are unable to capture nonlinear and bursty mobility patterns. 
Existing spatio-temporal models exhibit limited capacity to represent heterogeneous interactions and offer limited support for personalization. 
Consequently, their performance varies between datasets and prediction horizons.

\subsubsection{Cold-Start App and Location Recommendation}
We compare our approach with a diverse set of representative baselines including (i) GCN-based model \textbf{LLM-Init}~\cite{Zhang2025LLMInitAF}, aggregating user's historical prompt as user embedding. (2) SSM-based model \textbf{SIGMA}~\cite{Liu2024SIGMASG}, introducing MAMBA blocks to enhance the context modeling in sequence recommendation. (iii)
LLM-enhanced methods, \textbf{LLM-Emb}~\cite{Liu2024LLMEmbLL} and \textbf{LLM-ESR}~\cite{Liu2024LLMESRLL}, which utilize the user pattern extracted from LLM to fine-tune the main model.

In this cold start scenario, the objective is to recommend items in a target domain where the user lacks historical interactions.
We use the behavioral sequence from a related source domain over the previous 20 time points and then predict which items that will achieve the highest engagement frequency in subsequent time periods.
Conventional baselines address this challenge by learning latent structural correlations between the observed variables. 
To further enhance inference capabilities, LLM-enhanced approaches and our proposed U-MASK framework explicitly incorporate textual modality information and other user’s historical data with comprehensive multi-modal history from semantically similar users, enabling inference beyond direct interaction history.
For app recommendation, the model integrates the target user’s historical location with the rich network traffic, app, and location traces of semantically similar users. 
We conduct a symmetrical evaluation for the location recommendation task. Performance is assessed using Recall, NDCG, and MRR, adapting the cutoffs (@1, @3, @5) to the differing cardinalities of the app and location domains.


As shown in Tables~\ref{tab:app}-~\ref{tab:location}, U-MASK consistently outperforms GCN-based, SSM-based, and LLM-enhanced baselines under a unified protocol in which no external user profiles are provided to non-LLM methods, ensuring a fair comparison. 
This confirms that its advantage comes from modeling design rather than information asymmetry. 
For app recommendation, U-MASK reaches a Recall@5 of 0.9816, improving over the strongest LLM-enhanced BERT4Rec variant by 1.8\%. 
For location recommendation, it achieves Recall@1 of 0.9412 and NDCG@3 of 0.9706, surpassing the strong SSM-based SIGMA model. 
Notably, U-MASK operates without pre-existing user prompts or language models at inference time. 
Instead, it builds user representations in real time from sparse behavioral traces via U-SCOPE and applies task-aware masking to focus inference on informative spatio-temporal contexts

\begin{table}[t]
\centering
\caption{App Recommendation with User Profile. \textbf{Bold} values indicate the best performance.}
\label{tab:app}
\resizebox{\linewidth}{!}{%
\begin{tabular}{l l c c c c c c}
\toprule
\rowcolor{headergray}
 &  & \multicolumn{3}{c}{\textbf{@3}} & \multicolumn{3}{c}{\textbf{@5}} \\
\cmidrule(lr){3-5} \cmidrule(lr){6-8}
\rowcolor{headergray}
\multirow{-2}{*}{\textbf{Category}} & \multirow{-2}{*}{\textbf{Model}} 
& R & N & M & R & N & M \\
\midrule

GCN & LLM-Init 
& 0.8396 & 0.5302 & 0.4205 & 0.8505 & 0.5345 & 0.4227 \\
\midrule

SSM & SIGMA 
& 0.3914 & 0.3463 & 0.3308 & 0.4419 & 0.3676 & 0.3429 \\
\midrule

\multirow{2}{*}{\shortstack[l]{LLM-enhanced\\SASRec}} 
& LLM-Emb 
& 0.8929 & 0.7912 & 0.8035 & 0.9285 & 0.8354 & 0.8035 \\
& LLM-ESR 
& 0.7857 & 0.7594 & 0.7500 & 0.9286 & 0.8162 & 0.7804 \\
\midrule

\multirow{2}{*}{\shortstack[l]{LLM-enhanced\\BERT4Rec}} 
& LLM-Emb 
& 0.8928 & 0.8222 & 0.7976 & 0.9643 & 0.8514 & 0.8125 \\
& LLM-ESR 
& 0.8214 & 0.7377 & 0.7083 & 0.8929 & 0.7669 & 0.7244 \\
\midrule

\multirow{2}{*}{\shortstack[l]{LLM-enhanced\\GRU4Rec}} 
& LLM-Emb 
& 0.8214 & 0.7508 & 0.7262 & 0.8571 & 0.7950 & 0.7738 \\
& LLM-ESR 
& 0.6429 & 0.5412 & 0.5060 & 0.7143 & 0.5704 & 0.5220 \\
\midrule

\rowcolor{oursbg}
\textbf{Diffusion model} & \textbf{U-MASK} 
& \textbf{0.9601} & \textbf{0.9118} & \textbf{0.8824} & \textbf{0.9816} & \textbf{0.9236} & \textbf{0.9018} \\
\bottomrule
\end{tabular}%
}
\end{table}

\begin{table}[ht]
\centering
\caption{Location Recommendation with User Profile. \textbf{Bold} values indicate the best performance.}
\label{tab:location}
\resizebox{0.95\linewidth}{!}{%
\begin{tabular}{l l c c c c}
\toprule
\rowcolor{headergray}
 &  & \textbf{@1} & \multicolumn{3}{c}{\textbf{@3}} \\
\cmidrule(lr){3-3} \cmidrule(lr){4-6}
\rowcolor{headergray}
\multirow{-2}{*}{\textbf{Category}} & \multirow{-2}{*}{\textbf{Model}} 
& R & R & N & M \\
\midrule

GCN & LLM-Init 
& 0.8791 & 0.5912 & 0.6236 & 0.4945 \\
\midrule

SSM & SIGMA 
& 0.8977 & 0.9071 & 0.9161 & 0.9144 \\
\midrule

\multirow{2}{*}{\shortstack[l]{LLM-enhanced\\SASRec}} 
& LLM-Emb 
& 0.9285 & 0.9642 & 0.9511 & 0.9464 \\
& LLM-ESR 
& 0.6428 & 0.7500 & 0.6747 & 0.6488 \\
\midrule

\multirow{2}{*}{\shortstack[l]{LLM-enhanced\\BERT4Rec}} 
& LLM-Emb 
& 0.8571 & 0.9285 & 0.9153 & 0.9107 \\
& LLM-ESR 
& 0.7956 & 0.7857 & 0.6577 & 0.6130 \\
\midrule

\multirow{2}{*}{\shortstack[l]{LLM-enhanced\\GRU4Rec}} 
& LLM-Emb 
& 0.9285 & 0.9642 & 0.9511 & 0.9464 \\
& LLM-ESR 
& 0.4642 & 0.8571 & 0.7074 & 0.6547 \\
\midrule

\rowcolor{oursbg}
\textbf{Diffusion model} & \textbf{U-MASK} 
& \textbf{0.9412} & \textbf{0.9717} & \textbf{0.9706} & \textbf{0.9608} \\
\bottomrule
\end{tabular}%
}
\end{table}

\subsection{RQ2: Performance of U-MASK against Conventional Masking}
\label{subsec:rq7}


We evaluate U-MASK against standard spatio-temporal masking strategies under identical experimental conditions. 
The control variant shares the same backbone  architecture and input features as U-MASK. 
The only difference is that it uses a fixed masking strategy, while ours is adaptive.

Across seven real-world datasets, U-MASK consistently surpasses the standard masking approach in both short- and long-term multi-task prediction.
Relative performance gains range from approximately 5\% to over 90\% in prediction accuracy across evaluation settings.. 
The largest improvements are observed in location prediction. This task shows strong behavioral differences between users. The result highlights U-MASK's ability to find and focus on task-specific spatio-temporal features, thereby significantly enhancing prediction robustness.

\begin{table*}[t]
\centering
\caption{Comparison of short-term and long-term multi-task forecasting performance between traditional spatio-temporal masking (baseline) and our U-MASK.
\textbf{Bold} values indicate the best performance.
\textbf{\textcolor{posdelta}{Green}} values in $\Delta$ rows indicate relative improvement.}
\label{tab:multitask_polished}

\begingroup
\setlength{\tabcolsep}{2.0pt}        
\renewcommand{\arraystretch}{0.86}   
\setlength{\aboverulesep}{0pt}       
\setlength{\belowrulesep}{0pt}
\tiny

\newcommand{\ourscell}[1]{\cellcolor{oursbg}\textbf{#1}}

\resizebox{0.9\textwidth}{!}{%
\begin{tabular}{@{}l l c c c c c c c c c c c c@{}}
\toprule
\rowcolor{headergray}
& &
\multicolumn{6}{c}{\textbf{Short-term prediction}} &
\multicolumn{6}{c}{\textbf{Long-term prediction}} \\
\cmidrule(lr){3-8}\cmidrule(lr){9-14}
\rowcolor{headergray}
& &
\multicolumn{3}{c}{\textbf{RMSE}} & \multicolumn{3}{c}{\textbf{MAE}} &
\multicolumn{3}{c}{\textbf{RMSE}} & \multicolumn{3}{c}{\textbf{MAE}} \\
\cmidrule(lr){3-5}\cmidrule(lr){6-8}\cmidrule(lr){9-11}\cmidrule(lr){12-14}
\rowcolor{headergray}
\textbf{Dataset} & \textbf{Method} &
\multicolumn{1}{c}{APP} & \multicolumn{1}{c}{Traffic} & \multicolumn{1}{c}{Location} &
\multicolumn{1}{c}{APP} & \multicolumn{1}{c}{Traffic} & \multicolumn{1}{c}{Location} &
\multicolumn{1}{c}{APP} & \multicolumn{1}{c}{Traffic} & \multicolumn{1}{c}{Location} &
\multicolumn{1}{c}{APP} & \multicolumn{1}{c}{Traffic} & \multicolumn{1}{c}{Location} \\
\midrule

\multirow{3}{*}{\textbf{Dataset1}}
& Baseline & 0.0203 & 0.0054 & 0.0232 & 0.0074 & 0.0009 & 0.0096 & 0.0218 & 0.0066 & 0.0287 & 0.0107 & 0.0046 & 0.0184 \\
& \cellcolor{oursbg}\textbf{Ours}
& \ourscell{0.0129} & \ourscell{0.0042} & \ourscell{0.0145}
& \ourscell{0.0040} & \ourscell{0.0006} & \ourscell{0.0041}
& \ourscell{0.0201} & \ourscell{0.0054} & \ourscell{0.0241}
& \ourscell{0.0073} & \ourscell{0.0010} & \ourscell{0.0114} \\
& {$\Delta$}
& \textcolor{posdelta}{+36.4\%} & \textcolor{posdelta}{+22.2\%} & \textcolor{posdelta}{+37.5\%}
& \textcolor{posdelta}{+45.9\%} & \textcolor{posdelta}{+33.3\%} & \textcolor{posdelta}{+57.3\%}
& \textcolor{posdelta}{+7.8\%}  & \textcolor{posdelta}{+18.2\%} & \textcolor{posdelta}{+16.0\%}
& \textcolor{posdelta}{+31.8\%} & \textcolor{posdelta}{+78.3\%} & \textcolor{posdelta}{+38.0\%} \\
\midrule

\multirow{3}{*}{\textbf{Dataset2}}
& Baseline & 0.0141 & 0.0056 & 0.0418 & 0.0039 & 0.0115 & 0.0155 & 0.0154 & 0.0080 & 0.0393 & 0.0076 & 0.0063 & 0.0189 \\
& \cellcolor{oursbg}\textbf{Ours}
& \ourscell{0.0076} & \ourscell{0.0047} & \ourscell{0.0224}
& \ourscell{0.0028} & \ourscell{0.0028} & \ourscell{0.0070}
& \ourscell{0.0138} & \ourscell{0.0070} & \ourscell{0.0390}
& \ourscell{0.0041} & \ourscell{0.0052} & \ourscell{0.0171} \\
& {$\Delta$}
& \textcolor{posdelta}{+46.1\%} & \textcolor{posdelta}{+16.1\%} & \textcolor{posdelta}{+46.4\%}
& \textcolor{posdelta}{+28.5\%} & \textcolor{posdelta}{+75.7\%} & \textcolor{posdelta}{+54.8\%}
& \textcolor{posdelta}{+10.4\%} & \textcolor{posdelta}{+12.5\%} & \textcolor{posdelta}{+0.8\%}
& \textcolor{posdelta}{+46.1\%} & \textcolor{posdelta}{+17.5\%} & \textcolor{posdelta}{+9.5\%} \\
\midrule

\multirow{3}{*}{\textbf{Dataset3}}
& Baseline & 0.0206 & 0.0082 & 0.0164 & 0.0089 & 0.0034 & 0.0078 & 0.0247 & 0.0134 & 0.0188 & 0.0161 & 0.0099 & 0.0121 \\
& \cellcolor{oursbg}\textbf{Ours}
& \ourscell{0.0098} & \ourscell{0.0023} & \ourscell{0.0097}
& \ourscell{0.0025} & \ourscell{0.0010} & \ourscell{0.0028}
& \ourscell{0.0158} & \ourscell{0.0043} & \ourscell{0.0166}
& \ourscell{0.0052} & \ourscell{0.0008} & \ourscell{0.0069} \\
& {$\Delta$}
& \textcolor{posdelta}{+52.4\%} & \textcolor{posdelta}{+71.9\%} & \textcolor{posdelta}{+40.9\%}
& \textcolor{posdelta}{+71.9\%} & \textcolor{posdelta}{+70.7\%} & \textcolor{posdelta}{+64.1\%}
& \textcolor{posdelta}{+36.0\%} & \textcolor{posdelta}{+67.9\%} & \textcolor{posdelta}{+11.7\%}
& \textcolor{posdelta}{+67.7\%} & \textcolor{posdelta}{+91.9\%} & \textcolor{posdelta}{+43.0\%} \\
\midrule

\multirow{3}{*}{\textbf{Dataset4}}
& Baseline & 0.0184 & 0.0047 & 0.0167 & 0.0081 & 0.0009 & 0.0061 & 0.0171 & 0.0047 & 0.0181 & 0.0077 & 0.0091 & 0.0084 \\
& \cellcolor{oursbg}\textbf{Ours}
& \ourscell{0.0098} & \ourscell{0.0024} & \ourscell{0.0090}
& \ourscell{0.0031} & \ourscell{0.0005} & \ourscell{0.0028}
& \ourscell{0.0166} & \ourscell{0.0047} & \ourscell{0.0161}
& \ourscell{0.0070} & \ourscell{0.0021} & \ourscell{0.0061} \\
& {$\Delta$}
& \textcolor{posdelta}{+46.7\%} & \textcolor{posdelta}{+48.9\%} & \textcolor{posdelta}{+46.1\%}
& \textcolor{posdelta}{+61.7\%} & \textcolor{posdelta}{+44.4\%} & \textcolor{posdelta}{+54.1\%}
& \textcolor{posdelta}{+2.9\%}  & {0.0\%} & \textcolor{posdelta}{+11.0\%}
& \textcolor{posdelta}{+9.1\%}  & \textcolor{posdelta}{+76.9\%} & \textcolor{posdelta}{+27.4\%} \\
\midrule

\multirow{3}{*}{\textbf{Dataset5}}
& Baseline & 0.0191 & 0.0049 & 0.0172 & 0.0109 & 0.0022 & 0.0067 & 0.0271 & 0.0062 & 0.0187 & 0.0180 & 0.0042 & 0.0106 \\
& \cellcolor{oursbg}\textbf{Ours}
& \ourscell{0.0106} & \ourscell{0.0034} & \ourscell{0.0105}
& \ourscell{0.0029} & \ourscell{0.0011} & \ourscell{0.0031}
& \ourscell{0.0172} & \ourscell{0.0048} & \ourscell{0.0177}
& \ourscell{0.0061} & \ourscell{0.0013} & \ourscell{0.0070} \\
& {$\Delta$}
& \textcolor{posdelta}{+44.5\%} & \textcolor{posdelta}{+30.6\%} & \textcolor{posdelta}{+39.0\%}
& \textcolor{posdelta}{+73.4\%} & \textcolor{posdelta}{+50.0\%} & \textcolor{posdelta}{+53.7\%}
& \textcolor{posdelta}{+36.5\%} & \textcolor{posdelta}{+22.6\%} & \textcolor{posdelta}{+5.3\%}
& \textcolor{posdelta}{+66.1\%} & \textcolor{posdelta}{+69.0\%} & \textcolor{posdelta}{+34.0\%} \\
\midrule

\multirow{3}{*}{\textbf{Dataset6}}
& Baseline & 0.0192 & 0.0056 & 0.0217 & 0.0073 & 0.0012 & 0.0087 & 0.0250 & 0.0085 & 0.0223 & 0.0159 & 0.0066 & 0.0117 \\
& \cellcolor{oursbg}\textbf{Ours}
& \ourscell{0.0096} & \ourscell{0.0029} & \ourscell{0.0098}
& \ourscell{0.0028} & \ourscell{0.0005} & \ourscell{0.0030}
& \ourscell{0.0172} & \ourscell{0.0056} & \ourscell{0.0219}
& \ourscell{0.0121} & \ourscell{0.0012} & \ourscell{0.0100} \\
& {$\Delta$}
& \textcolor{posdelta}{+50.0\%} & \textcolor{posdelta}{+48.2\%} & \textcolor{posdelta}{+54.8\%}
& \textcolor{posdelta}{+61.6\%} & \textcolor{posdelta}{+58.3\%} & \textcolor{posdelta}{+65.5\%}
& \textcolor{posdelta}{+31.2\%} & \textcolor{posdelta}{+34.1\%} & \textcolor{posdelta}{+1.8\%}
& \textcolor{posdelta}{+23.9\%} & \textcolor{posdelta}{+81.8\%} & \textcolor{posdelta}{+14.5\%} \\
\midrule

\multirow{3}{*}{\textbf{Dataset7}}
& Baseline & 0.0211 & 0.0061 & 0.0153 & 0.0115 & 0.0038 & 0.0079 & 0.0257 & 0.0079 & 0.0174 & 0.0149 & 0.0056 & 0.0113 \\
& \cellcolor{oursbg}\textbf{Ours}
& \ourscell{0.0113} & \ourscell{0.0040} & \ourscell{0.0091}
& \ourscell{0.0039} & \ourscell{0.0019} & \ourscell{0.0032}
& \ourscell{0.0168} & \ourscell{0.0054} & \ourscell{0.0162}
& \ourscell{0.0066} & \ourscell{0.0016} & \ourscell{0.0077} \\
& {$\Delta$}
& \textcolor{posdelta}{+46.4\%} & \textcolor{posdelta}{+34.4\%} & \textcolor{posdelta}{+40.5\%}
& \textcolor{posdelta}{+66.1\%} & \textcolor{posdelta}{+50.0\%} & \textcolor{posdelta}{+59.5\%}
& \textcolor{posdelta}{+34.6\%} & \textcolor{posdelta}{+31.6\%} & \textcolor{posdelta}{+6.9\%}
& \textcolor{posdelta}{+55.7\%} & \textcolor{posdelta}{+71.4\%} & \textcolor{posdelta}{+31.9\%} \\
\bottomrule
\end{tabular}%
}
\endgroup
\end{table*}

\subsection{RQ3: Overall Performance of Data Generation}

To evaluate whether key behavioral relationships are preserved in the U-SCOPE's generated data, we analyze the correlations between app usage and location types. In particular, we examine whether the generated dataset still preserves the essential app-location patterns found in the original data~\cite{yu2018smartphone}.

\begin{figure*}[t]
\centering
\includegraphics[width=0.92\linewidth]{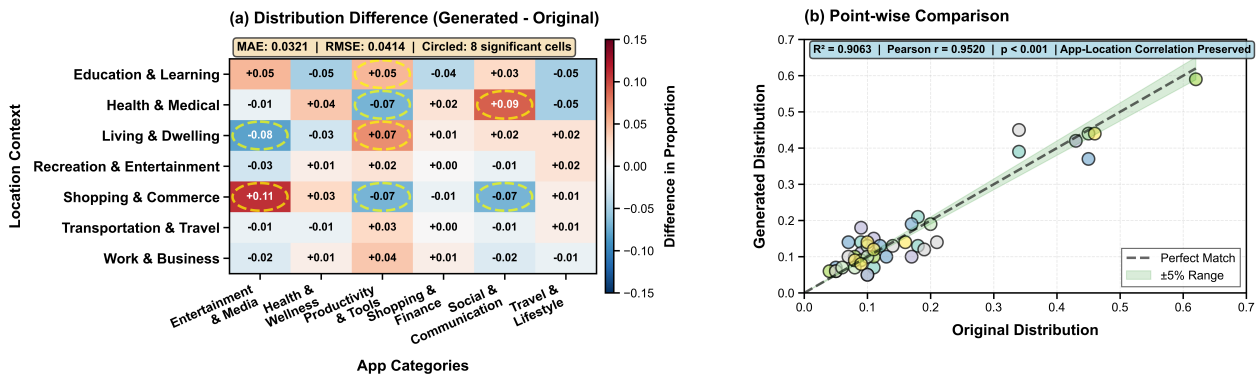}
\caption{Comparison of app usage distributions
(a) Difference heatmap (Generated -- Original) with significant deviations 
($\lvert \mathrm{diff} \rvert > 0.05$) circled.
(b) Point-wise comparison showing strong correlation 
($R^{2} = 0.9063$, $p < 0.001$) between original and generated data, 
indicating preservation of app--location relationships.
}
\label{fig_heatmaps}
\end{figure*}

As illustrated in Fig.~\ref{fig_heatmaps}, the generated data demonstrates a high degree of fidelity in replicating the complex app-location correlations present in the original data. 
Fig.~\ref{fig_heatmaps}(a) shows the distribution difference heatmap where both the MAE is of 0.0457 and the RMSE of RMSE=0.0543 indicate minimal deviation across all location contexts. 
For instance, the dominant usage of "Entertainment \& Media" app in "Recreation \& Entertainment" contexts is accurately preserved. 
Furthermore, the point-wise comparison in Fig.~\ref{fig_heatmaps}(b) provides statistical evidence of this alignment. The scatter plot reveals a clear linear relationship between the original and generated distributions, with a coefficient of determination ($R^2$) of 0.9063 and a Pearson correlation coefficient ($r$) of 0.9520 ($p < 0.001$). 
The majority of data points fall within the $\pm 5\%$ tolerance range shown in green area, confirming that the generation framework effectively maintains fundamental app-location relationships without introducing significant artifacts. 



\subsection{RQ4: User Profile Embedding Evaluation}
To evaluate the utility of user profiles generated by U-SCOPE, we incorporate them into all baseline models employed in the U-MASK evaluation.
We augment each baseline with a conditional integration module. Specifically, the U-SCOPE profile is processed by a semantic encoder and subsequently fed into the model as an auxiliary contextual input. 
As shown in Tables~\ref{tab1}–\ref{tab6}, this enhancement consistently improves performance across all models on multivariate prediction tasks.
These results indicate that although U-SCOPE embeddings are derived from anonymous interaction data, they encode transferable behavioral patterns that benefit diverse baseline architectures.
\begin{figure*}[t]
\centering
\includegraphics[width=0.95\linewidth]{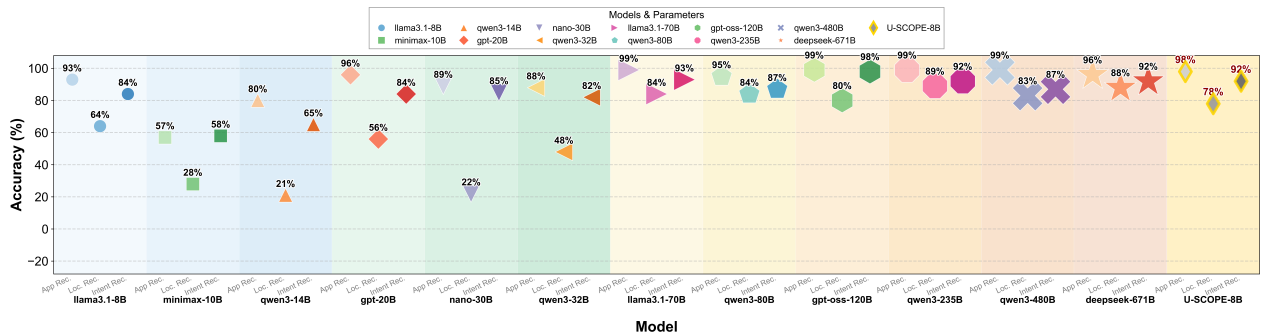}
\caption{Performance comparison of different models on cold-start recommendations.}
\label{fig_recom}
\end{figure*}

\subsection{RQ5: Fine-tuned LLM Evolution}

We examine whether combining synthetic behavioral data with user profiles can improve the adaptability of LLMs in recommendation scenarios. 
In particular, we fine-tune a pre-trained LLM using generated user data and evaluate its capability to infer preferences for previously unseen app, locations, or intents.
The training procedure follows a reasoning-guided knowledge distillation strategy. We first leverage a frontier model (e.g., Qwen3-235B) to generate explicit Chain-of-Thought (CoT) trajectories for tasks on which it exhibits superior performance. These high-quality reasoning traces are then incorporated into the fine-tuning dataset for a second-stage supervised fine-tuning on an 8B backbone. By learning these structured reasoning patterns, the smaller model internalizes the teacher model's heuristic capabilities, allowing its performance to closely approximate that of frontier models.
\begin{figure}[t]  
    \centering
    \includegraphics[width=\linewidth]{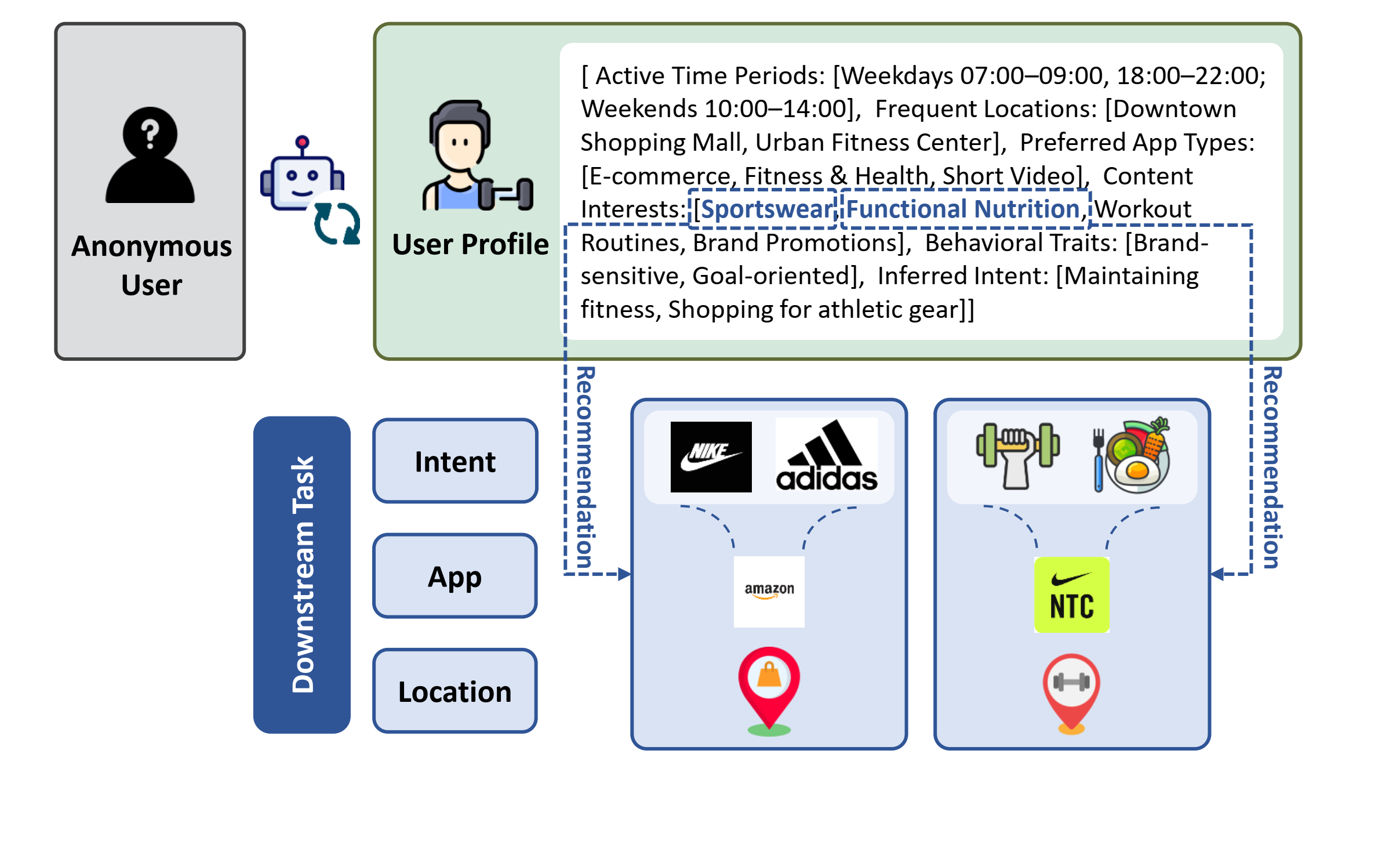}  
    \caption{U-SCOPE can build user profiles for anonymous users and uses these profiles to provide personalized recommendations for apps, locations, and intents.}
    \label{uscope_intro}
\end{figure}

Specifically, the base model is \textbf{Llama-3.1-8b} ~\cite{touvron2023llama}, fine-tuned using \textbf{LoRA} for parameter-efficient adaptation~\cite{hu2022lora}. The teacher model is \textbf{Qwen3-235B} ~\cite{yang2025qwen3}. Experiments are conducted on a single NVIDIA A100 GPU. Fine-tuning on 500 users and CoT trajectories requires approximately 1.5 hours. 
As illustrated in Fig.~\ref{uscope_intro}, U-SCOPE generates user profiles for anonymous users and leverages them to enable personalized recommendations of apps, locations, and intents. 
The novelty recommendation scenario aims to predict user preferences for new apps, locations, or intents that have never appeared in the user’s past interactions. It achieves this by fusing historical behavior with the semantic U-SCOPE user profile, enabling generalization to previously unseen contexts.

We compare U-SCOPE against language models spanning a wide range of scales, from 8B to over 480B parameters.
As shown in Fig.~\ref{fig_recom}, U-SCOPE achieves an average accuracy of 89.33\%, significantly outperforming models such as Qwen3-14B at 55.33\%, Minimax-M2\cite{chen2025minimax} at 47.67\%, Nano-30B at 65.33\%, and Qwen3-32B at 72.67\%. Remarkably, this compact model obtained through efficient fine-tuning on synthetic user data reaches performance close to that of the largest systems including Llama3.1-70b, DeepSeek-v3\cite{liu2025deepseek}, GPT-OSS-120B\cite{gptoss_2023}, and Qwen3-235B. On individual tasks, U-SCOPE attains 98\% for app recommendation, 78\% for location recommendation, and 92\% for intent prediction.
This demonstrates that with efficient fine-tuning on synthetic data, smaller models can achieve performance close to that of much larger models, particularly in cold-start scenarios where new, unseen user contexts are prevalent. 
The success of U-SCOPE highlights the importance of leveraging synthetic user profiles and advanced fine-tuning techniques to optimize recommendation performance, even in cases where larger models may not adapt effectively.
These findings provide strong evidence that our method, despite relying on a lightweight model, can deliver high-quality, personalized recommendations in cold-start settings.


\begin{figure*}[t]
    \centering
    \includegraphics[width=1.0\linewidth]{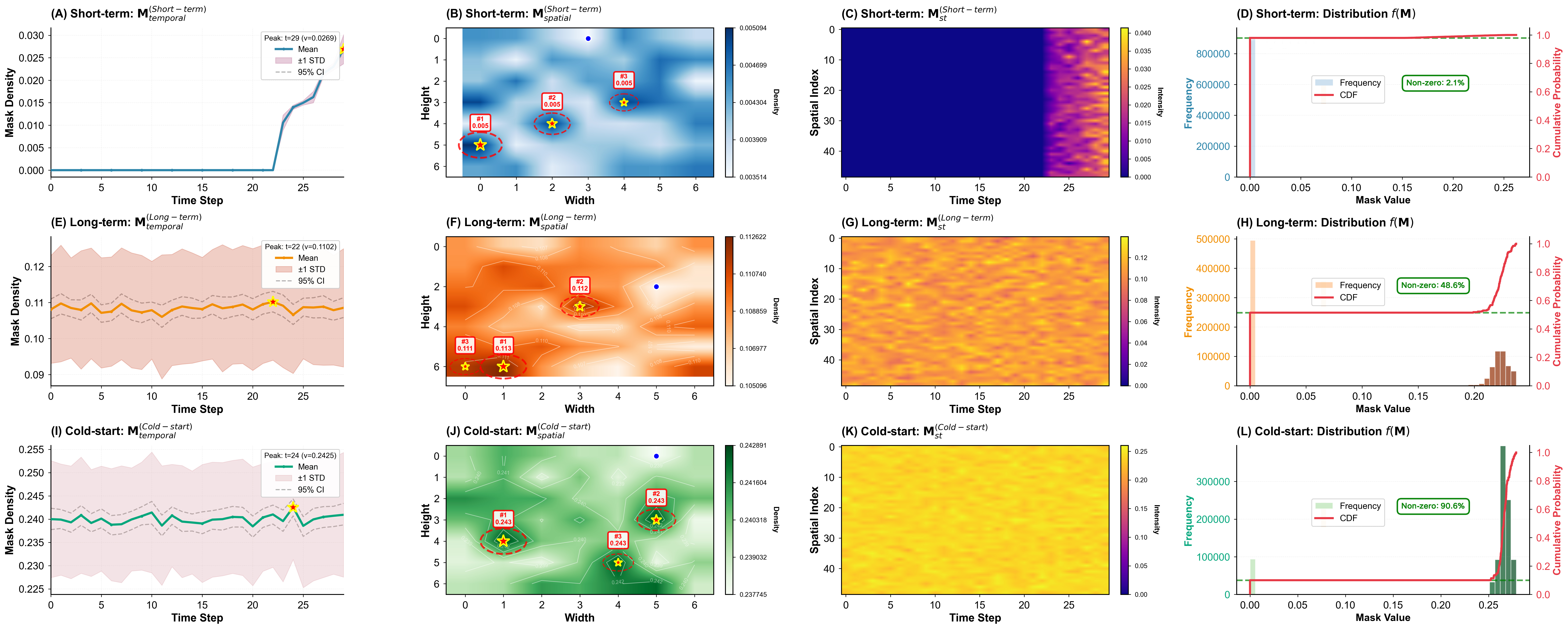}
     \caption{\textbf{Task-granularity-aware masking through task-specific feature adaptation.}
    U-MASK learns to modulate spatio-temporal feature selection in response to task demands via a lightweight, task-specific feature refinement module. The $3 \times 4$ grid visualizes this adaptivity: rows correspond to \textbf{short-term prediction}, \textbf{long-term forecasting}, and \textbf{cold-start modeling}; columns show \textbf{(1) temporal masks}, \textbf{(2) spatial masks}, \textbf{(3) joint spatio-temporal masks}, and \textbf{(4) mask value distributions}. Across tasks the model automatically shifts from sparse localized masking in short term prediction to dense exploratory coverage in cold start scenarios. This behavior reveals a self organized information bottleneck that balances specificity and robustness without any manual intervention.}
    \label{fig:mask_dist}
\end{figure*}

\subsection{RQ6: Task-Granularity-Driven Spatio-Temporal Masking}
\label{subsec:rq6}
We analyze the spatio-temporal masks learned by U-MASK across three standard tasks, including short-term prediction, long-term forecasting, and cold-start modeling, using real-world mobile interaction data from 128 users, as shown in Fig.~\ref{fig:mask_dist}. 
All masking patterns are produced by our lightweight task adaptation module (Eq.~\ref{eq:refinement_umask}), without any task-specific hyperparameter tuning. This enables direct analysis of how the task adaptation model tailors spatio-temporal feature selection to each task's granularity.

Fig.~\ref{fig:mask_dist} visualizes these masks, which are produced by the meta-adaptation module without manual hyperparameter tuning.
In short-term prediction, recent interactions are highly informative, so the model uses sparse masks that emphasize the latest time steps and a localized spatial region. 
In long-term prediction, behavior evolves dynamically, prompting the model to widen its temporal receptive field with moderately dense masks that balance historical coverage and noise suppression, while maintaining selective but broader spatial attention. 
In cold-start recommendation, where user history is unavailable, the model adopts a dense exploratory mode, distributing attention nearly uniformly over the spatio-temporal grid to maximize information extraction from limited observations.
These results confirm that the task adaptation model dynamically tailors feature selection to the specific granularity and information reliability of each task.

A key point is that these different masking strategies provide empirical evidence for our theoretical framework. Sparse local masks capture Immediacy. Extended temporal windows maintain Stability. Dense exploratory attention enables Generalization.
U-MASK successfully unifies these conflicting requirements within the Impossibility Triangle through a single adaptive mechanism.

\subsection{RQ7: Behavior-Driven Personalization of Masking Strategies}
\label{subsec:rq2}

\begin{figure*}[t]
    \centering
    \includegraphics[width=1.0\linewidth]{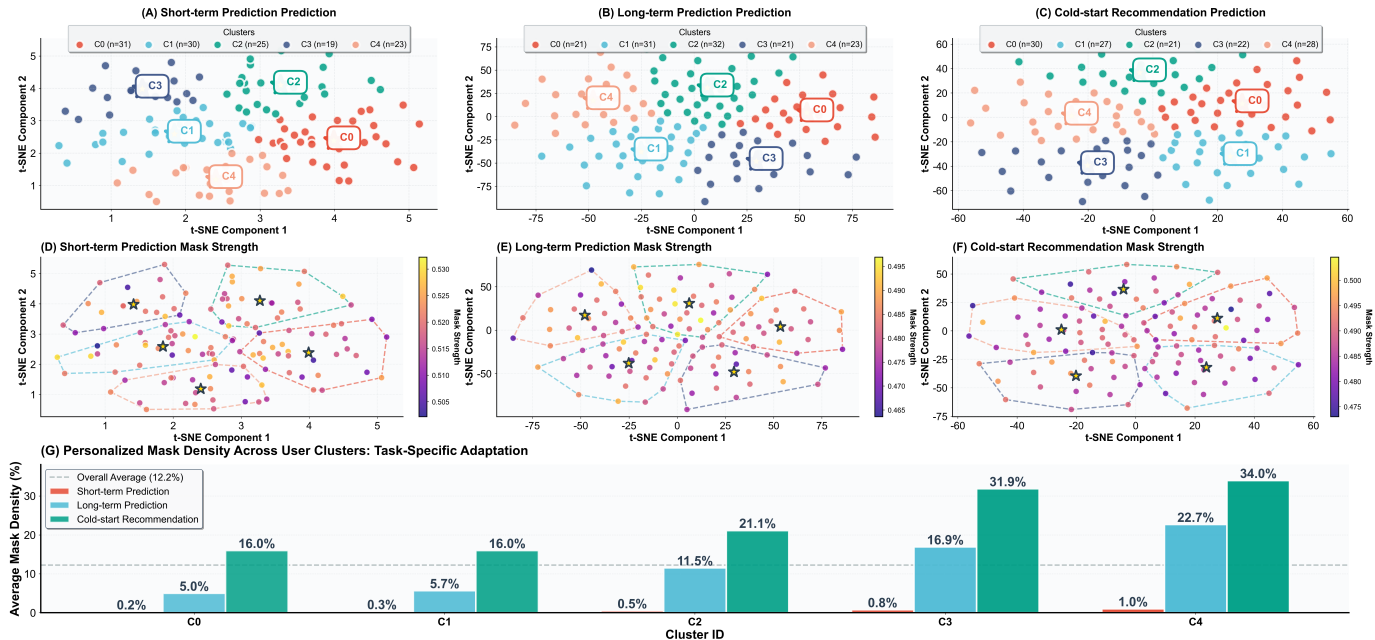}
    \caption{\textbf{Personalized masking through behavior-aware clustering.}
        \textbf{Row 1 (Behavioral Clustering)}: t-SNE projection of U-MASK's pattern features reveals five distinct user clusters (C0–C4), indicating that users naturally segment into interpretable behavioral groups.
        \textbf{Row 2 (Cluster-Conditioned Masking}: The spatial distribution of mask strengths aligns closely with cluster boundaries, demonstrating that masking strategies are conditioned on latent behavioral patterns.
        \textbf{Row 3 (Personalized Sparsity Patterns)}: Mask density varies systematically across clusters, reflecting how U-MASK tailors feature selection to individual predictability and behavioral complexity.}
        \label{fig:clustering}

\end{figure*}

To answer this question, we analyze how U-MASK leverages latent behavioral patterns to derive a personalized masking policy for each user. As shown in Fig.~\ref{fig:clustering}, we perform task-aware clustering on the learned pattern features $\mathbf{F}^{\text{pat}}$ using K-Means ($k=5$) in the t-SNE embedded space.
This analysis identifies five distinct user groups with corresponding masking strategies. 
For instance, users in Cluster C0 exhibit highly regular behavior, such as consistent daily commutes. 
Consequently, U-MASK assigns sparse masks, since minimal context is sufficient for accurate inference.
In contrast, Cluster C4 display more diverse and multi-context behavior. U-MASK therefore assigns denser masks to capture the increased uncertainty and variability in their activity patterns.

As illustrated in Row 2 of Fig.~\ref{fig:clustering}, the mask strength distributions correspond closely to the cluster boundaries. Furthermore, mask density transitions smoothly across the feature space. 
This indicates that personalization arises from a general policy learned from individual dynamics instead of from fixed rules tied to specific clusters.
By directly connecting masking strategies to latent behavioral patterns, U-MASK achieves true personalization. 
It adapts feature selection to each user's predictability and behavioral complexity, showing that personalized masking arises from learned representations of individual dynamics.

\section{Conclusion}\label{con}
This work examined a fundamental challenge in personalized mobile AI, namely the structural tension among immediacy, stability, and generalization when inferring user behavior from sparse histories under dynamic spatio-temporal context. 
By formulating personalized behavior modeling as conditional completion on a partially observed spatio-temporal behavior tensor, we introduced U-MASK, a user-adaptive spatio-temporal masking approach that unifies short-term prediction, long-horizon forecasting, and cold-start inference within a single framework. Through task-specific and user-adaptive mask generation, U-MASK enables accurate and robust inference in complex and data-scarce settings. 
Extensive experiments on real-world mobile datasets demonstrated consistent improvements over existing methods across considered scenarios, particularly under severe data sparsity. 
An important direction for future work is to incorporate richer multimodal spatio-temporal signals, such as satellite imagery, urban street views, and structured behavioral data, to further enhance contextual understanding and generalization in personalized mobile AI systems.


\bibliographystyle{IEEEtran}
\bibliography{references}

\begin{IEEEbiographynophoto}{Shiyuan Zhang} is a Ph.D. student at the Department of Electrical and Electronic Engineering, The University of Hong Kong. His research interests include generative AI, computer networks, and mobile computing.
\end{IEEEbiographynophoto}
\begin{IEEEbiographynophoto}{Yilai Liu} is an MPhil student at the Department of Electrical and Electronic Engineering, The University of Hong Kong. His research focuses on generative models and large language models, with a particular interest in temporal and multimodal learning.
\end{IEEEbiographynophoto}
\begin{IEEEbiographynophoto}{Yuwei Du} is a Master's candidate in the Department of Electronic Engineering at Tsinghua University, specializing in spatial intelligence, large language models, human behavior modeling, urban science, and spatiotemporal data mining. 
\end{IEEEbiographynophoto}
\begin{IEEEbiographynophoto}{Ruoxuan Yang} is a Ph.D. student at the Department of Electrical and Electronic Engineering, The University of Hong Kong. Her research interests include human-computer interaction and future-oriented prototyping.
\end{IEEEbiographynophoto}
\begin{IEEEbiographynophoto}{Hongyang Du} is an assistant professor at the Department of Electrical and Electronic Engineering, The University of Hong Kong, where he directs the Network Intelligence and Computing Ecosystem (NICE) Laboratory. He received the Ph.D. degree from the Nanyang Technological University, Singapore. His research interests include edge intelligence, generative AI, and network management.
\end{IEEEbiographynophoto}
\begin{IEEEbiographynophoto}{Dong In Kim}
(Fellow, IEEE) is a Distinguished Professor with the College of Information and Communication Engineering, Sungkyunkwan University, Suwon, South Korea. He received the Ph.D. degree in electrical engineering from the University of Southern California, Los Angeles, CA, USA. He is a Fellow of the Korean Academy of Science and Technology and a Member of the National Academy of Engineering of Korea.
\end{IEEEbiographynophoto}

\vfill
\end{document}